\documentclass{article}

\usepackage[preprint]{neurips_2022}

\usepackage[utf8]{inputenc} 
\usepackage[T1]{fontenc}    
\usepackage{hyperref}       
\usepackage{url}            
\usepackage{booktabs}       
\usepackage{amsfonts}       
\usepackage{nicefrac}       
\usepackage{microtype}      
\usepackage{xcolor}         

\usepackage{bm}
\usepackage{amsmath}
\definecolor{mygray}{gray}{0.6}
\definecolor{cadmiumgreen}{rgb}{0.0, 0.42, 0.24}
\usepackage{siunitx}
\usepackage{graphicx}
\usepackage[normalem]{ulem}
\usepackage{algorithm}
\usepackage{algorithmic}

\title{Adversarial Example Detection\\in Deployed Tree Ensembles}

\author{%
Laurens Devos\\
KU Leuven\\
\texttt{laurens.devos@kuleuven.be}
\And
Wannes Meert\\
KU Leuven\\
\texttt{wannes.meert@kuleuven.be}
\And
Jesse Davis\\
KU Leuven\\
\texttt{jesse.davis@kuleuven.be}
}

\begin{document}

\maketitle

\begin{abstract}
 Tree ensembles are powerful models that are widely used. However, they are susceptible to adversarial examples, which are examples that purposely constructed to elicit a misprediction from the model. This can degrade performance and erode a user's trust in the model. Typically, approaches try to alleviate this problem by verifying how robust a learned ensemble is or robustifying the learning process. We take an alternative approach and attempt to detect adversarial examples in a post-deployment setting. We present a novel method for this task that works by analyzing an unseen example's output configuration, which is the set of predictions  made by an ensemble's constituent trees. Our approach works with any additive tree ensemble and does not require training a separate model.  We  evaluate  our  approach  on three different tree ensemble learners. We empirically show that our method is currently the best adversarial detection method for tree ensembles.
 
\end{abstract}

\section{Introduction}

Tree ensembles such as (gradient) boosted trees and random forests are a popular class of models. However, like many other model families, such as neural networks \cite{szegedy2014,goodfellow2015,biggio2013}, they are susceptible to adversarial attacks \cite{kantchelian2016,einziger2019,hchen19,zhang2020,wang2020lp,devos21}. While there are many types of attacks, this paper explores the setting where the model is deployed and operating in the wild, and exposed to adversarial examples. Intuitively in this context an adversarial example is one that is purposely constructed to elicit a misprediction from the model~\cite{diochnos18}. 
Such examples are undesirable because they degrade a model's performance and erode a user's trust in the model. For tree ensembles, the literature attempts to deal with this in one of two ways. First, verification techniques attempt to ascertain how robust a learned ensemble is to adversarial examples~\cite{hchen19,ranzato2020abstract,devos21}. This is often done by empirically determining how much an example would have to be perturbed (according to some norm) for its predicted label to change. Second, the problem can be addressed at training time by trying to learn a more robust model by adding adversarial examples to the training set~\cite{kantchelian2016}, pruning the training data~\cite{yang_rashtchian20b}, changing aspects of the learner such as the splitting criteria~\cite{hchen2019robust,andriushchenko2019,vos2021efficient}, or interleaving learning and verification~\cite{ranzato2021genetic}.

This paper explores an alternative approach to mitigating the effect of adversarial examples in a post deployment setting. Given a current example for which a prediction is required, we attempt to ascertain if this current example is adversarial or not. If the example is identified as being adversarial, then the deployed model could refrain from making a prediction similar to a learning with rejection setting~\cite{Cortes2016ALT}.
While this question has been extensively explored for neural networks, this is not the case for tree ensembles. Unfortunately, most existing methods for neural networks are not applicable to tree ensembles because they use properties unique to neural networks~\cite{ZhangShigeng2022}. For example, some modify the model ~\cite{grosse2017,gong2017,metzen2017}, learn other models (e.g., nearest neighbors) on top of the network's intermediate representations~\cite{feinman2017,lee2018simple,katzir2019,sperl2020dla}, or learn other models on top of the gradients~\cite{schulze2021da3g}.
Moreover, nearly all methods focus on detecting adversarial examples only in the context of image classification.

Tree ensembles are powerful because they combine the predictions made by many trees. Hence, the prediction procedure involves sorting the given example to a leaf node in each tree. The ordered set of the reached leaf nodes is an \textit{output configuration} of the ensemble and fully determines the ensemble's resulting prediction.
However, there are many more possible output configurations than there are examples in the data used to train the model. For example, the California housing dataset~\cite{pace1997} only has eight features, but training an XGBoost ensemble containing 6, 7, or 8 trees each of at most depth 5 yields 62\,248, 173\,826, and 385\,214 output configurations respectively.\footnote{Computed using Veritas~\cite{devos21}.} These numbers (far) exceed the 20,600 examples in the dataset. The situation will be worse for the larger ensembles sizes that are used in practice. Our hypothesis is that 
\begin{quote}
    adversarial examples exploit {\bf unusual output configurations}, that is, ones that are very different to those observed in the data used to train the model.
\end{quote}
That is, small, but carefully selected perturbations can yield an example that is quite similar to another example observed during training, but yields an output configuration that is far away from those covered by the data used to train the model. 

Based on this intuition, we present a novel method to detect adversarial examples based on assessing whether an example encountered post deployment has an unusual output configuration.
When an example is encountered post deployment, our approach encodes it by its output configuration and then measures the distance between the encoded example and its nearest (encoded) neighbor in a reference set.
If this distance is sufficiently high, the example is flagged as being an adversarial one and the model can abstain from making a prediction. Our approach has several benefits. First, it is general: it works with any additive tree ensemble. Second, it is  integrated: it does not require training a separate model to identify adversarial examples, one simply has to set a threshold on the distance. Finally, it is surprisingly fast as the considered distance metric can be efficiently computed by exploiting instruction level parallelism (SIMD).  

Empirically, we evaluate and compare our approach on three ensemble methods: gradient boosted trees (XGBoost~\cite{chen2016xgboost}), random forests \cite{breiman2001random}, and GROOT~\cite{vos2021efficient}, which is a recent approach for training robust tree ensembles.
We empirically show that our method outperforms multiple competing approaches for detecting adversarial examples post deployment for all three considered tree ensembles. Moreover, it can detect adversarial examples with a comparable computational effort.

\section{Preliminaries}

Given an input space $\mathcal{X}$ and an output space $\mathcal{Y}$,
we detect adversarial examples in deployed models learned from a dataset $D \subseteq \mathcal{X}\times\mathcal{Y}$ mapping $\mathcal{X}$ to $\mathcal{Y}$.

\subsection{Additive tree ensembles}

This paper proposes a method that works with additive ensembles of decision trees (e.g., those available in XGBoost~\cite{chen2016xgboost}, LightGBM~\cite{ke2017lightgbm} and Scikit-learn~\cite{scikit-learn}). This encompasses both random forests and (gradient) boosted decision trees.

A binary tree is a recursive data structure consisting of nodes. It has one root node, which is the only node that is not a descendent of any other node. Every node is either a leaf node containing an output value, or an internal node storing a test (e.g., \textit{is $X_i$ less than 5?}) that indicates whether to go left or right, and references to left and right sub-trees.
A decision tree $T$ is evaluated on an example $x \in \mathcal{X}$ by starting at the root and executing the tests until a leaf node is reached.

An additive ensemble of decision trees is a sum of trees. The prediction $\bm{T}(x)$ for an example $x \in \mathcal{X}$ involves summing up the predicted leaf values of each decision tree in the ensemble: $\sum_m T_m(x)$, $m=1, \ldots, M$, with $M$ the number of trees in the ensemble.

\subsection{Output configurations}

\begin{figure}
    \centering
    \footnotesize
    \includegraphics{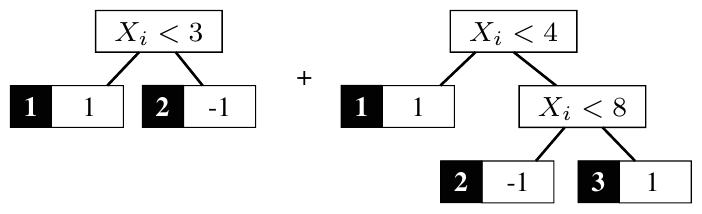}
    \caption{A simple tree ensemble with in each leaf an identifier in black, and leaf class predictions}
    \label{fig:trees_example}
\end{figure}

The output configuration (OC) of an example $x$ is the ordered set of leaf nodes $(l_1, \ldots, l_M)$ that are reached when evaluating the trees in the ensemble.
An output configuration corresponds to a feasible combination of root-to-leaf paths where there is one such path for each tree in the ensemble.
We define a mapping $\mathrm{OC} : x \mapsto (l_1, \ldots, l_M)$ mapping an example $x$ to its output configuration. We call the discrete space of all feasible output configurations $\mathcal{O} = \{ \mathrm{OC}(x) \mid x \in\mathcal{X} \}$ the \textbf{OC-space} of an ensemble.
This OC-space $\mathcal{O}$ corresponds to equivalence classes in \cite{tornblom2020}.
Note that $\mathcal{O}$ is not just the Cartesian product of all leaves. This is because some combinations of leaves are invalid, e.g., in Figure~\ref{fig:trees_example}, $(1, 2)$ and $(1, 3)$ are invalid configurations because $X_i$ cannot be both less than 3 and greater than 4 at the same time.

\subsection{Adversarial examples}

We use the same definition of adversarial examples as \cite{kantchelian2016,hchen19,devos21} and others: $\tilde{x}$ is an adversarial example of $x \in \mathcal{X}$ when three conditions hold: (1) $\|\tilde{x}-x\|$ is small according to some norm, (2) $\bm{T}(x)$ returns the correct label for $x$, and (3) $\bm{T}(\tilde{x}) \neq \bm{T}(x)$.
This is similar to the \textit{prediction-change} setting of \cite{diochnos18}, which only requires conditions (1) and (3).

\section{Detecting adversarial examples}

We assume a post-deployment setting where a tree ensemble $\bm{T}$ is operating in the wild. Our task can then be defined as follows:
\begin{description}
\item[Given:]  a deployed tree ensemble $\bm{T}$ and an example $x \in \mathcal{X}$ for which a prediction is required
\item[Do:] assign a score to $x$ indicating whether $x$ is an adversarial example
\end{description}

Our algorithm is based on the fact that for sufficiently large models, \textbf{the vast majority of the model's possible outputs configurations will not be observed in the data used to train the model.} Given this insight, our hypotheses are that

 \begin{enumerate}
     \item[\textbf{H1.}] 
Adversarial examples arise because certain minimal changes to the original example can produce large changes in the output configuration, causing the model's prediction to change.
     \item[\textbf{H2.}]
Adversarial examples produce highly \textit{unusual} output configurations, exploiting the fact that most such configurations were never observed in the data used to train the model.
 \end{enumerate}

Decision tree learners employ heuristics to select a split criterion in each internal node that helps distinguish among the different classes. Consequently, most leaf nodes tend to be (strongly) predictive of one class.
In an ensemble, correctly classified positive examples  will tend to have output configurations consisting largely of leaves that predict the positive class whereas the converse is true for negative examples. 
Adversarial examples contain carefully selected perturbations to a small number of feature values that result in more leaves of the opposing class appearing in an  output configuration, yielding  an \textit{unusual} output configuration with an incorrect output label. 
This suggests that measuring how similar a newly encountered example's output configuration is to those that appear in the training set will be an effective way to detect adversarial examples.

First, we discuss how to measure the distance between output configurations. Second, we introduce our \textit{OC-score} metric which computes the distance between a newly encountered example and the reference set. A higher score indicates that the example is more likely adversarial. Finally, we show how to efficiently compute the \textit{OC-score} using instruction-level parallelism.

\subsection{Distances between output configurations}

Following the intuition that adversarial examples exploit unusual output configurations, our method measures the abnormality of an output configuration by comparing it to the typical output configurations traversed by examples in a reference set.
Two output configurations $O = \{l_1, \ldots, l_M\}$ and $O' = \{ l'_1, \ldots, l'_M \}$ can be compared using the Hamming distance:
\begin{equation}
\label{eq:ham}
\mathrm{ham}(O, O') = \sum_{m=1}^M I(l_m \neq l'_m).
\end{equation}
The Hamming distance counts the number of leaves that differ between the two output configurations. It measures the distance between two examples \textit{in OC-space} rather than the input space $\mathcal{X}$. This is isomorphic to the proximity metric in Random Forests~\cite{breiman:rfmanual}.

\subsection{The \textit{OC-score} metric: distance to the closest reference set example}
Our approach requires a learned ensemble $\bm{T}$ and a subset $D_R$ of the data $D$ used to train the model. It constructs a reference set $R = \{ \mathrm{OC}(x) \mid x \in D_R \}$ by encoding the examples in $D_R$ into the OC-space by finding each one's output configuration.
In practice, $R$ is a matrix of small integers (e.g. \texttt{uint8}) that identify the leaves (e.g. the black identifiers in Figure~\ref{fig:trees_example}). The $i$th row in $R$ contains the identifiers of the leaves in the output configuration of the $i$th example in $D_R$
In the experiments, we take $R$ to be the output configurations of the correctly classified training examples.

Given a newly encountered example $x \in \mathcal{X}$, the ensemble is used to obtain its predicted label $\hat{y}=\bm{T}(x)$ and output configuration $\mathrm{OC}(x)$.
Then it receives a \textit{OC-score} by computing the OC-space distance to the closest example in a reference dataset $R$:
\begin{equation}
    \operatorname{OC-score}(x) = \min_{O' \in R[y=\hat{y}]} \mathrm{ham}(\mathrm{OC}(x), O'),
    \label{eq:pscore}
\end{equation}
where $R[y=\hat{y}]$ is the subset of examples in $R$ with label $\hat{y}$. Higher \textit{OC-scores} correspond to a higher chance of being an adversarial example.
To operationalize this, a threshold can be set on the \textit{OC-scores} to flag potential adversarial examples: when the threshold is exceeded the model should abstain from making a prediction.

\subsection{Fast distance computations via SIMD}

Finding an example's nearest neighbors in the OC-space can be done very efficiently by exploiting the fact that ensembles tend to produce trees with a relatively small number of leaf nodes. Hence it is possible to assign each leaf in a tree a short binary code word that can be represented by a small integer and exploit instruction-level parallelism using SIMD to compute the Hamming distance.

To compute $\operatorname{OC-score}(x)$, we need to compute the Hamming distance to each example in the reference set, that is, we need to slide the vector $\mathrm{OC}(x)$ over the rows of $R$. The trees used in the experiments have no more than 256 leaves. Hence, when using the 256-bit \texttt{AVX2} registers, we can compute the Hamming distance of 32 reference set examples in parallel. This massively speeds-up the computation even when $R$ is large. The SIMD pseudo-code is given in the supplement.

\section{Related Work}
Beyond the approaches mentioned in the introduction for detecting adversarial examples in neural networks, there are methods that look at the behavior of the decision boundary in an example's neighborhood \cite{fawzi2018adversarial,roth2019,tian2022detecting}. Unfortunately, these methods do not work well with tree ensembles because the use of binary axis-parallel splits make them step functions, which makes it difficult to extract information from the neighborhood. Also, relevant to this paper is the work investigating the relation between model uncertainty and adversarial examples~\cite{liu2018advbnn,grosse2018limitations}.

The random forest manual~\cite{breiman:rfmanual} discusses defining distances between training examples in an analogous manner to \textit{OC-score}. Typically, (variations on) this distance has been used for tasks such as clustering~\cite{shi:rf-cluster2006} or making tree ensembles more interpretable~\cite{tan2020tree}. To our knowledge, it has not been used for detecting adversarial examples. 
Alternatively, some works propose different ways to use the path an example follows in a tree ensemble to reencode them, typically with the idea of using this encoding as a new feature space for training a model~\cite{vens:icdm11,pliakos:2016}.

Each example's \textit{OC-score} can be viewed as a model's secondary output with the predicted class being its  primary output. This fits into the larger task of machine learning with a reject option \cite{Cortes2016ALT}.
Rejection aims to identify test examples for which the model was not properly trained. For such examples, the model's predictions have an elevated risk of being incorrect, and hence may not be trustworthy. An example can be rejected due to ambiguity (i.e., how well the decision boundary is defined in a region) or novelty (i.e., how anomalous an example is wrt the observed training data)~\cite{Hendrickx2021arXiv}. The \textit{OC-score} metric goes beyond measuring ambiguity in an ensemble (i.e., the model's confidence in a prediction). Therefore, it can detect adversarial examples even if they fall in a region of the input space where the model's decision boundary appears to be well defined given the training data.

\section{Experimental Evaluation}
Our experimental evaluation addresses three questions:\footnote{The supplement addresses a fourth question: How does the proportion of adversarial examples in the test set affect performance of our \textit{OC-score} metric?} 
\begin{enumerate}
    \item[Q1.] Can our approach more accurately detect adversarial examples than its competitors?
    \item[Q2.] What is each approach's prediction time cost associated with detecting adversarial examples?
    \item[Q3.] How does the size of the reference set affect the performance of our \textit{OC-score} metric? 
\end{enumerate}

We compare our \textit{OC-score} to four approaches:
    
    \textbf{Ambiguity (\textit{ambig})} This approach uses the intuition that because adversarial examples are somehow different than the training ones, the model will be uncertain about an adversarial example's predicted label~\cite{grosse2018limitations}.
    This entails deciding whether an example lies near a model's decision boundary. This can be done by ranking examples according to the uncertainty of the classifier:
    \begin{equation}
    \textrm{ambig}(x) = 1 - |2\,p_{\bm{T}}(x)-1|,
    \end{equation}
    where $p_{\bm{T}}$ is the probability of the positive class as predicted by the ensemble $\bm{T}$ for an example $x$.
    
\textbf{Local outlier factor (\textit{lof})} \cite{lof} Another intuition to detect adversarial examples is to employ an anomaly detector under the assumption that adversarial examples are drawn from a different distribution than non-adversarial ones.  \textit{lof} is a state-of-the-art unsupervised anomaly detection method that assigns a score to each example denoting how anomalous it is. This approach entails learning a \textit{lof} model which is  applied to each example. 

\textbf{Isolation forests (\textit{iforest})}
        An isolation forest~\cite{iforest} is state-of-the-art anomaly detector. It learns a tree ensemble that separates anomalous from normal data points by splitting on a randomly selected attribute using a randomly chosen split value between the minimum and maximum value of the attribute.
    Outliers tend to be split off earlier in the trees, so the depth of an example in the tree is indicative of how  normal an example is.    Again, this requires learning a separate model at training time. 
    
    \paragraph{ML-LOO}  This is an approach for detecting adversarial examples from the neural network literature \cite{yang2020mlloo}. Unlike most other approaches, it is model agnostic as it looks at statistics of the features. It uses the accumulated \textit{feature attributions} to rank examples:
    \begin{equation}
        \mathrm{mlloo}(x) = \mathrm{std}_k\left\{p_{\bm{T}}(x) - p_{\bm{T}}(x_{(k)})\right\},
    \end{equation}
    where $p_{\bm{T}}$ is the probability prediction of ensemble $\bm{T}$, and $x_{(k)}$ is $x$ with the $k$th attribute set to 0.
    The observation in \cite{yang2020mlloo} is that variation in the feature attributions is larger for adversarial examples.

\begin{table}[htbp]
    \centering
    \small
    \caption{Datasets characteristics and learners' hyperparameter settings. The  characteristics are the   number of features \#F and examples  $n$.  The learning rate for XGBoost is $\eta$.  Each ensemble contains $M$ trees and $d$ is the maximum tree depth. }
    \begin{tabular}{lccccc}
        \toprule
        \textbf{Name}& \#F &$n$ & $\eta$ &$M$ &$d$  \\
        \midrule
        phoneme      &   5 &  5.4k & .5 & 20 & 6    \\
        spambase     &  57 &  4.6k & .5 & 20 & 5    \\
        covtype      &  54 &  581k & .5 & 80 & 6    \\
        higgs        &  33 &  250k & .1 & 100& 8    \\
        ijcnn1       &  22 &  142k & .9 & 50 & 5    \\
        mnist2v4     & 784 &  13.8k& .7 & 50 & 5    \\
        fmnist2v4    & 784 &  14k  & .9 & 50 & 5    \\
        webspam      & 254 &  350k & .9 & 50 & 5    \\
        \bottomrule
    \end{tabular}
    \label{tab:datasets}
\end{table}

\subsection{Experimental methodology}
We mimic the post-deployment setting using 5-fold cross validation. In each fold, a model is trained on clean training data (4 folds). Then, in the remaining fold some of the correctly classified examples are perturbed and turned into adversarial examples. This fold augmented by the perturbed examples is used as the unseen examples. The task is then, given the model and a reference set $R$, to detect which of the examples in the unseen set are adversarial using the \textit{OC-score}.

We test our approach on the eight benchmark datasets listed in Table~\ref{tab:datasets}.
All datasets are min-max normalized to make perturbations of the same size to different attributes comparable. To demonstrate our approach's generality, we consider three types of additive tree ensembles: (1) XGBoost boosted trees \cite{chen2016xgboost}, (2) Scikit-learn random forests \cite{scikit-learn}, and (3)  GROOT robustified random forests~\cite{vos2021efficient}, which modifies the criteria for selecting the split condition when learning a tree to make them more robust against adversarial examples. Due to space constraints, we only show plots for 4 out of 8 datasets. The results for the remaining datasets are along the same lines and are provided in the supplement.
 
{\bf Experimental settings}
 For a given dataset, each learner has the same number of trees in the ensemble and each tree is restricted to be of the same maximum depth. Details for each dataset are given in Table~\ref{tab:datasets}.
We use the scikit-learn~\cite{scikit-learn} implementation for \textit{lof} and \textit{iforest} and use the default scikit-learn hyper-parameters. 
The supplement reports the average accuracies of the learned models on each dataset and the attack model $\epsilon$ of the GROOT ensembles.
All experiments ran on an Intel E3-1225 with 32GB of memory. Multi-threading was enabled for all methods.

{\bf Generating adversarial examples} We generate adversarial examples using Veritas~\cite{devos21} with the $l_\infty$ norm.
Per fold in the benchmarks, we generate three different sets of adversarial examples. Each set is based on 500 randomly selected, correctly classified test set examples.
The first set are the closest adversarial examples. For each of these adversarial examples, it is guaranteed that no other adversarial example exists that is closer to the original example.
This set of examples corresponds to the ones generated by the Kantchelian et al.'s MILP approach \cite{kantchelian2016}.
The second set of adversarial examples allows perturbations of size $2\cdot \delta_{\mathrm{med}}$, where $\delta_{\mathrm{med}}$ is the median adversarial perturbation observed in the set of closest adversarial examples.
The third set has a larger perturbations of size $5\cdot \delta_{\mathrm{med}}$.
We refer to these three sets as \textit{closest adv.}, \textit{adv. x2}, and \textit{adv. x5}. Each set of adversarial examples thus has its own properties. The closest adversarial examples tend to barely cross a model's decision boundary whereas for the adversarial examples in \textit{adv. x5} the models could return extremely confident outputs. The supplement contains illustrative adversarial examples for the image datasets \textit{mnist2v4} and \textit{fmnist2v4}. For each set of 500 adversarial examples, we construct the final evaluation set by adding 2500 randomly selected normal previously unseen (i.e., not used to train the model) test set examples, apart from phoneme and spambase where we select 1080 and 920 normal examples, respectively.\footnote{Because these two datasets have fewer examples.}

\subsection{Results Q1: detecting adversarial examples}

The task is to distinguish the 500 adversarial examples from the normal test set examples. We measure detection performance in two different ways:  by evaluating ranking and  coverage versus detection rate tradeoff. 

\begin{figure}
    \centering
    \footnotesize
    \includegraphics{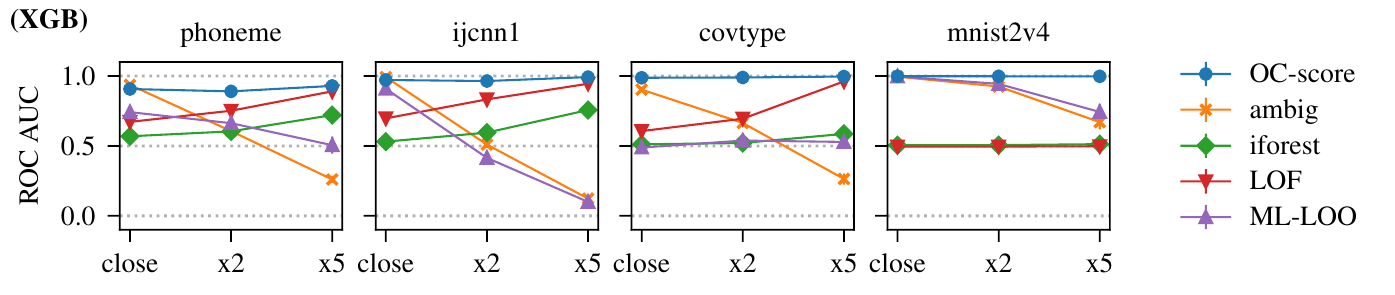}
    
    \vspace{0.5em}
    
    \includegraphics{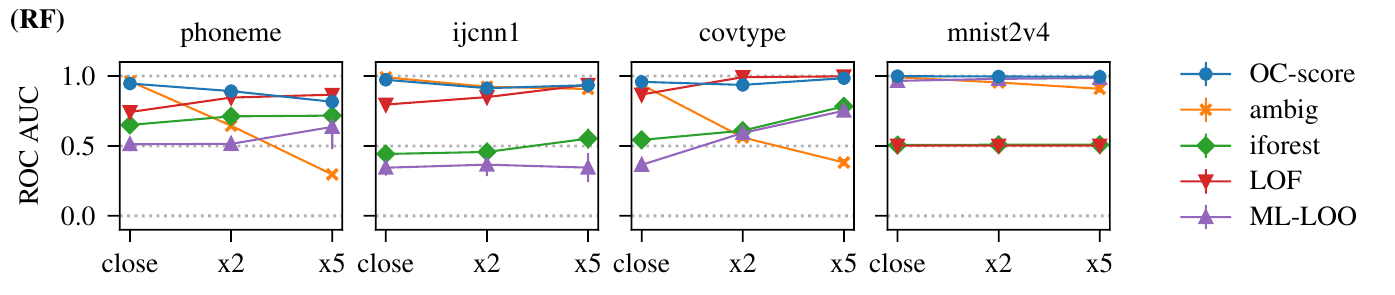}
    
    \vspace{0.5em}
    
    \includegraphics{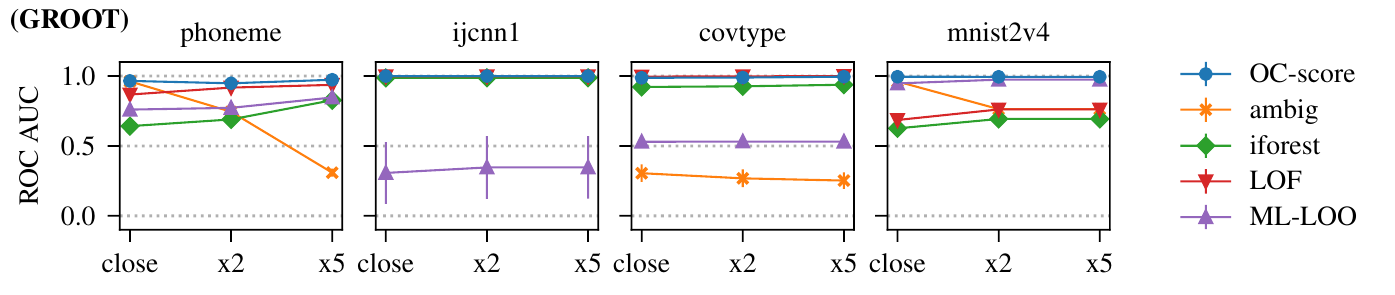}

    \caption{The average AUC ROC values for detecting adversarial vs. non-adversarial test examples as a function of the size of the adversarial perturbation for our score (\textit{OC-score}), ambiguity (\textit{ambig}), isolation forest (\textit{iforest}), local outlier factor (\textit{lof}), and ML-LOO for each of the adversarial sets. The results are shown for 4 out of 8 datasets for each of the three considered tree ensemble learners. The results for the other 4 datasets are similar and are in the supplement.}
    \label{fig:auc_small}
\end{figure}

\begin{figure*}[h!]
    \centering
    \footnotesize
    \includegraphics{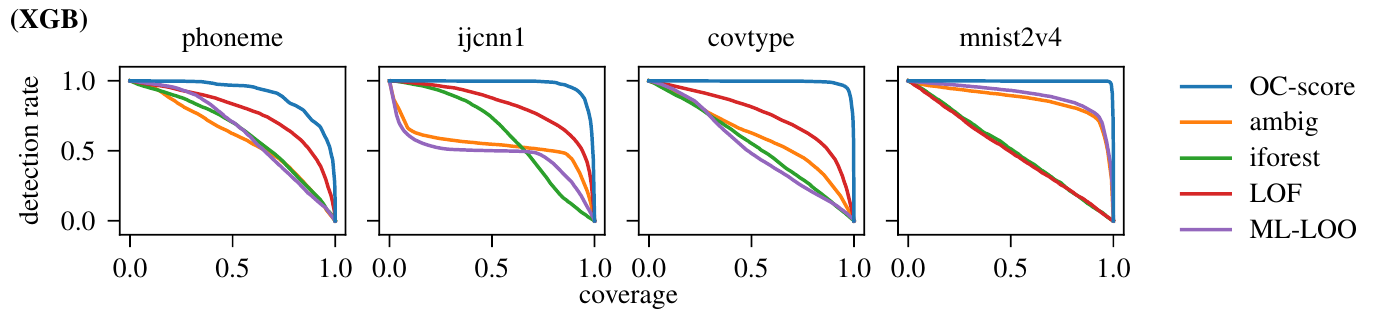}
    
    \vspace{0.5em}
    
    \includegraphics{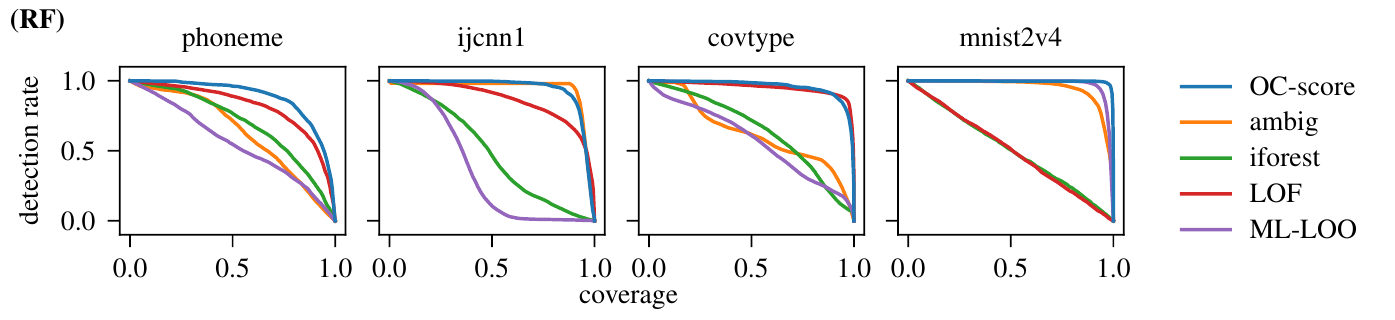}
    
    \vspace{0.5em}
    
    \includegraphics{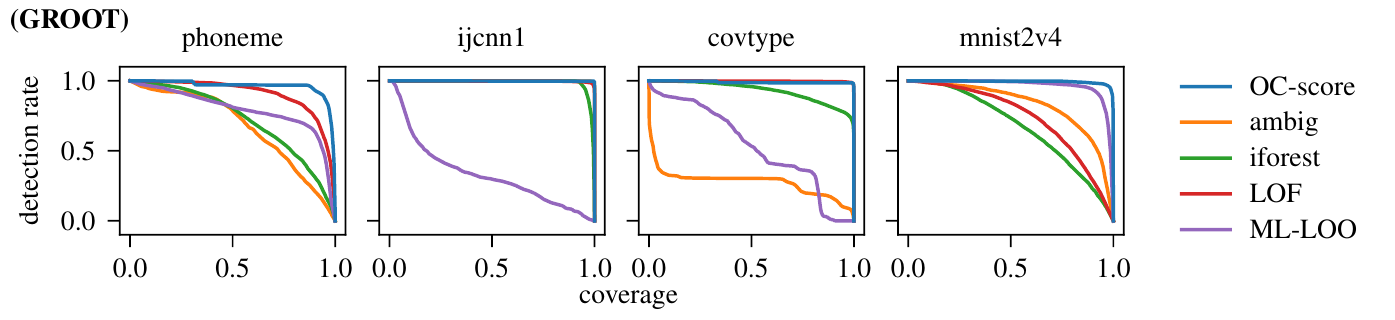}
    
    \caption{Coverage versus detection rate of adversarial examples for each method using three ensemble types. The results are averaged over the three sets of adversarial examples. The higher the curve, the better. The results for the remaining datasets are similar and in the supplement.}
    \label{fig:detection_rate}
\end{figure*}

{\bf Ranking.}
Each method assigns every test example a score, which can be used to rank examples from most to least likely that they are adversarial. The area under the ROC (AUC ROC) curve measures the quality of a ranking with respect to the classification task of separating adversarial from normal examples. In this case, it captures an approach's ability to distinguish adversarial examples from non-adversarial ones.  
Figure~\ref{fig:auc_small} shows the AUC ROC as a function of the magnitude of the adversarial perturbation for all methods, datasets, and ensemble learners.

Our \textit{OC-score} consistently performs the best overall. Its  performance is stable across the three considered types of adversarial examples for all three  ensemble learning techniques.

Ambiguity is effective at detecting the closest adversarial examples on all datasets and ensemble types. This is unsurprising as the closest adversarial examples tend to be constructed by perturbing the attribute values just enough to cause the generated adversarial example to fall just on the other side of a model's decision boundary. Hence, by definition, the model is uncertain for these adversarial examples. Ambiguity's ability to detect adversarial examples declines when they exhibit larger perturbations. However, an exception is  \textit{ijcnn1} and \textit{mnist2v4} for Random Forests and GROOT, where ambiguity detects most adversarial examples.
For \textit{phoneme}, the AUC values consistently drop below 0.5 for larger perturbation sizes. This arises because the model makes exceedingly confident incorrect predictions for these adversarial examples, which causes normal examples to be ranked as more adversarial than actual adversarial examples. To illustrate: the mean ambiguity score is 0.017 for normal examples and drops from 0.68  for \textit{close} adversarial examples to 0.008 for \textit{adv.x5} ones.

The anomaly detectors (\textit{iforest} and \textit{lof}) tend to perform poorly on most settings. The exceptions are the \textit{adv. x5} examples on \textit{phoneme}, \textit{covtype}, and \textit{ijcnn1} because these start to become  out of sample (i.e., very far away from the training data). These are also the datasets with the smallest number of attributes.
These methods also do better on GROOT where the modified split criteria produces more robust trees. Hence, the adversarial examples are farther away from the normal ones. 

 ML-LOO has highly variable performance and is consistently worse than \textit{OC-score}. It tends to work best on image data. However, for random forests and GROOT, it performs extremely poorly on several other datasets (e.g., \textit{ijcnn1}, \textit{covtype}). For the XGBoost ensembles, ML-LOO is generally effective at detecting the closest adversarial examples, but its performance degrades for the more distant ones.  

\begin{table*}[htbp]
    \centering
    \small
    \caption{Evaluation time results in milliseconds for the experiments displayed in Figure~\ref{fig:auc_small}. The evaluation time shows the average time to compute the score for a single test example. It is averaged over 5 folds, three ensemble types, and three sets of adversarial examples. }
    \begin{tabular}{lS[table-format=1.4]S[table-format=1.4]S[table-format=1.4]S[table-format=1.4]S[table-format=1.4]S[table-format=1.4]S[table-format=1.4]S[table-format=1.4]}
\toprule
{} & {phoneme} & {spambase} & {covtype} & {higgs} & {ijcnn1} & {mnist2v4} & {fmnist2v4} & {webspam} \\
\midrule
{OC-score}  &   0.00192 &     0.0019 &     0.973 &   0.587 &   0.0665 &     0.0221 &      0.0218 &     0.224 \\
{ambig}   &    0.0419 &     0.0314 &    0.0198 &  0.0202 &   0.0161 &     0.0278 &      0.0278 &    0.0203 \\
{iforest} &    0.0408 &     0.0461 &    0.0347 &  0.0348 &   0.0307 &      0.139 &       0.139 &      0.06 \\
{LOF}     &    0.0133 &     0.0677 &      7.23 &    3.06 &     1.84 &      0.307 &       0.308 &      5.55 \\
{ML-LOO}  &    0.0638 &      0.103 &      0.28 &   0.331 &     0.12 &       1.54 &        1.71 &     0.567 \\
\bottomrule
\end{tabular}

    \label{tab:timings}
\end{table*}

{\bf Coverage versus detection rate tradeoff.}
We now evaluate the detection approaches in an operational context where a model will first assess if an unseen example is adversarial or not. The model will only make predictions in cases where an examples is not flagged as being adversarial.  In practice, this requires thresholding the scores produced by each method to arrive at a hard decision as to whether an example is adversarial.  This induces a tradeoff between a method's (a) coverage, which is the fraction of normal test examples that are correctly identified as being normal and hence a prediction is made, and (b) detection rate, or the percent of correctly identified adversarial examples.
Figure~\ref{fig:detection_rate} shows each approach's detection rate as a function of coverage. The results average over all three types of adversarial examples. In nearly all cases, \textit{OC-score} results in a higher detection for each level of coverage than all its competitors. 
Figure~\ref{fig:detection_rate} only shows results for 4 out of 8 datasets, the results for the remaining datasets are in the supplement.

\subsection{Q2: Prediction time cost}

The run time cost has two parts: the setup time and the evaluation time.  The setup time is the one-time cost incurred during training by some of the methods. This is negligible for most methods, apart from \textit{lof} and is discussed in detail in the supplement. 

The evaluation time measures the overhead of computing a score indicating how likely it is that a test example is adversarial. Table~\ref{tab:timings} reports the average per example evaluation time for each method on every dataset. Regardless of the method, the scores can typically be computed well under 1 millisecond, with some cases taking slightly longer.
Unsurprisingly, \textit{ambiguity} is almost always the fastest as this is a simple mathematical computation.  \textit{iforest} is also fast because it only requires executing a tree ensemble. Computing the \textit{OC-score} is also relatively efficient due to exploiting SIMD.  It scales with number of trees and the size of the reference set, yielding higher evaluation times for large datasets and ensemble sizes (i.e., \textit{covtype} and \textit{higgs}). However, the results in Subsection~\ref{subsec:refset_size} indicate that it is possible to decrease the size of the reference set without degrading performance.  ML-LOO's costs comes from computing an importance for each feature. Hence, it yields longer times for datasets with many features (\textit{mnist2v4}, \textit{fmnist2v4}, \textit{webspam}). \textit{lof} is almost always the (second) slowest because computing the local density for each example is expensive.

\subsection{Q3: Effect of the reference set's size}
\label{subsec:refset_size}
\begin{figure}
    \centering
    \footnotesize
    \includegraphics{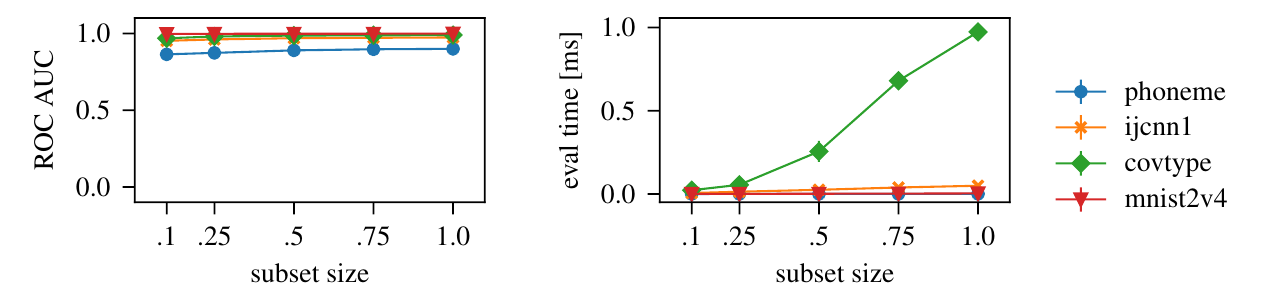}
    \caption{Left:  \textit{OC-score}'s ROC AUC values for detecting adversarial vs. non-adversarial test examples as a function of the reference set size. Right: average time in milliseconds to compute the \textit{OC-score} per example as a function of size reference set size.  Results are for XGBoost ensembles. Full results are in the supplement.}
    \label{fig:refsize}
\end{figure}

Finally, we explore the effect of the size of the reference set by varying the proportion of correctly classified training examples in the reference set. Figure~\ref{fig:refsize} shows the results for this experiment on four datasets using XGBoost ensembles. The plot on the left shows ROC AUC values for detecting adversarial vs. non-adversarial test examples. On all datasets, these values are relatively stable. There is a small decline in performance for the smallest reference set proportion, where the number of examples in the reference set ranges from 342 (\textit{spambase}, RF) to 41\,700 (\textit{covtype}, XGB).

The plot on the right shows the average time in milliseconds to compute the \textit{OC-score} per example as a function of the size of the reference set. The three smallest datasets show modest increases in the evaluation time for increasing reference set sizes. Higgs, which is by far the largest dataset, shows a steep increase as the size of the reference set increases which arises due to cache effects (i.e., misses) for the larger reference set sizes.
However, these experiments suggest that using a  small reference set would drastically improve the run time on this dataset without degrading performance. Note that changing the reference set does not require relearning the underlying ensemble used to make predictions, which is still learned using the full training set. 

\section{Conclusions and Discussion}
This paper explored how to detect adversarial examples post deployment for additive tree ensembles by detecting unusual output configurations. Our approach works with any additive tree ensemble and does not require training a separate model. 
If a newly encountered example's output configuration differs substantial from those in the training set, then it is more likely to be an adversarial example. Empirically,  our proposed \textit{OC-score} metric resulted in superior detection performance than existing approaches for three different tree ensemble learners on multiple benchmark datasets. One limitation of our work is that operationalizing it requires setting a threshold on the OC-score which may be difficult in practice. Moreover, while our approach achieves good performance it is unclear what constitutes acceptable detection performance for a deployed system because this is use-case dependent. While our approach makes it more difficult to perform adversarial attacks, it may also inspire novel strategies to construct  adversarial examples.

\section*{Acknowledgments}

This work was supported by iBOF/21/075, Research Foundation-Flanders (EOS No. 30992574, 1SB1320N to LD) and the Flemish Government under the ``Onderzoeksprogramma  Artifici\"ele  Intelligentie (AI) Vlaanderen'' program.


\bibliographystyle{plainnat}
\bibliography{ocscore}

\pagebreak

\appendix

\section{Fast \textit{OC-score} computation using SIMD}

Finding an example's nearest neighbors  in the OC-space can be done very efficiently by exploiting the fact that ensembles tend to produce trees with a relatively small number of leaf nodes. This is especially true for boosted ensembles like the ones generated by XGBoost~\cite{chen2016xgboost} and LightGBM~\cite{ke2017lightgbm}, where the number of leaves are often limited to less than 256 (e.g. maximum depth 8). Hence it is possible to assign each leaf in a tree a short binary code word (e.g. an 8-bit unsigned integer). An output configuration is then an array of $M$ short binary codes. This makes it possible to exploit instruction-level parallelism using SIMD to compute the Hamming distance between an output configuration of a new example, and the output configurations in the reference set $R$.

\begin{algorithm}[tb]
   \caption{Fast \textit{OC-score} computation for new example $x$}\label{alg:pscore}
\begin{algorithmic}[1]
    \STATE {\bfseries Input:} identifier vector $[\mathrm{id}_1, \ldots, \mathrm{id}_M]$ of $x$
    \STATE {\bfseries Input:} $|D_R| \times M$ col.maj. byte matrix $R$
    \STATE $\mathit{reg}_0$, $\mathit{reg}_1$, $\mathit{sum}$, and $\mathit{acc}$ are 32-byte registers
    \STATE initialize 32-byte register $\mathit{acc}$ to all 255s
    \STATE $\mathit{acc} \leftarrow \mathrm{setall}_v(255)$
    \FOR{$i=0 \ldots |D_R|$ {\bfseries in steps of} 32}
        \STATE $\mathit{sum} \leftarrow \mathrm{setall}_v(0)$
        \FOR{$m=0 \ldots M$}
            \STATE $\mathit{reg}_0 \leftarrow \mathrm{setall}_v(\mathrm{id}_m)$
            \STATE $\mathit{reg}_1 \leftarrow \mathrm{load}_v(R)[
                i\ldots (i+32), m])$
            \STATE $\mathit{sum} \leftarrow \mathrm{add}_v(\mathit{sum},
                \mathrm{and}_v(
                    \mathit{reg}_0 \neq_v \mathit{reg}_1, 1)
                    )$
        \ENDFOR
        \STATE $\mathit{acc} \leftarrow \mathrm{min}_v(\mathit{acc}, \mathit{sum})$
    \ENDFOR
    \STATE {\bfseries Return:} $\operatorname{OC-score}(x) = \mathrm{min}(\mathit{acc})$
\end{algorithmic}
\end{algorithm}

In the evaluation of this paper, we generate trees with at most 256 leafs. Hence, we can uniquely identify each leaf (per tree) with an 8-bit identifier.
For any leaf $l^{(m)}$ of some tree $T_m$, $m=1,\ldots,M$, let $\operatorname{id}_m(l^{(m)})$ be the 8-bit identifier of that leaf (e.g. a depth-first numbering of the leaves).
An output configuration $\operatorname{OC}(x)$ of an example $x$ is then an array of $M$ 8-bit identifiers:
\begin{equation}
    \operatorname{OC}(x) = [ \operatorname{id}_m(l^{(m)}) \mid \text{$x$ sorts to leaf $l^{(m)}$ in tree $T_m$}, m = 1,\ldots,M\ ]
\end{equation}
The reference set becomes an 8-bit matrix, with the output configurations of the reference set examples $x \in D_R$ in its rows:
\begin{equation}
    R = 
    \begin{bmatrix}
        \text{\----} & OC(x_1) & \text{\----} \\
                     & \vdots  &              \\
        \text{\----} & OC(x_{|D_R|}) & \text{\----}
    \end{bmatrix}
\end{equation}

To compute the OC-score of a new example $x$, we need to slide its identifier vector $\operatorname{OC}(x)$ over the rows of $R$, compute the Hamming distance between the identifier vector and the row, and keep track of the minimum distance. This can be done very quickly using SIMD. More specifically, we use the AVX2 extension to the x86 instruction set which introduces CPU instructions that operate on 32-byte wide registers, permitting computations on all 32 values simultaneously.

Algorithm~\ref{alg:pscore} shows how to use SIMD to compute the OC-score of a new example $x$.
We store $R$ in column-major order which means the values in the columns of $P$ are stored consecutively in memory. We assume that the number of examples in our reference set $|D_R|$ is a multiple of 32 for brevity.
The function $\mathrm{setall}_v(w)$ sets all 32 bytes in a register to the byte value $w$.
The other vectorized functions, indicated with a subscript $v$, perform the operations implied by their name on all 32 values simultaneously.
The $\mathrm{acc}$ register maintains 32 OC-score values, and is reduced to a single value at line 15.

Using unsigned bytes for the identifiers limits the ensemble's size in two ways. First, a byte only has 256 unique values, which restricts the number of leaves in each tree to be at most 256. Second, the sum on Line 11 could potentially overflow when the ensemble has more than 255 trees. Using a 16-bit short would alleviate this limitation,  but make the algorithm about two times slower.

\section{Predictive performance of used tree ensembles}\label{app:accuracies}

At the basis of each experiment in this paper are tree ensembles learned from data: we generate adversarial examples for these ensembles, and the OC-score uses the OC-space defined by the trees. We mimic the \textit{post-deployment} setting, which means that we consider the tree ensembles given, and are not concerned with the circumstances in which it was learned, i.e., the point of this paper is not to train the most accurate models, nor to deal with adversarial attacks during the training phase.
However, for completeness sake, we do include the test set accuracies and empirical robustness values achieved by the XGBoost, Random Forest and GROOT ensemble models to show that we are using competitive models.

Table~\ref{tab:accuracies} shows the average accuracy over five folds for each ensemble type on each dataset. XGBoost generally performs the best.
Random forests and GROOT ensembles tend to produce more robust models than XGBoost. They exhibit better empirical robustness scores as shown by the $\delta_{\mathrm{med}}$ values in Table~\ref{tab:median_delta} and the adversarial examples generated for these ensembles tend to be further away from the original distribution.  \textit{lof} and \textit{iforest}, which are particularly well suited at detecting if a test example is  out-of-sample, are better at detecting adversarial examples for these ensembles than ones for XGBoost. 

\begin{table}
    \begin{minipage}{.4\linewidth}
        \caption{The average accuracy for the learned models for XGBoost (XGB), Random Forests (RF), and GROOT.}%
        \label{tab:accuracies}
        \centering
        \footnotesize
        \begin{tabular}{llS[table-format=1.2]S[table-format=1.2]S[table-format=1.2]S[table-format=1.2]S[table-format=1.2]S[table-format=1.2]S[table-format=1.2]S[table-format=1.2]}
\toprule
{} & {XGB} &  {RF} & {GROOT} \\
\midrule
{phoneme}   &  0.88 &  0.86 &    0.79 \\
{spambase}  &  0.95 &  0.92 &    0.83 \\
{covtype}   &  0.89 &  0.76 &    0.69 \\
{higgs}     &  0.85 &  0.83 &    0.69 \\
{ijcnn1}    &  0.99 &  0.93 &    0.91 \\
{mnist2v4}  &  0.99 &  0.98 &    0.97 \\
{fmnist2v4} &  0.87 &  0.84 &    0.71 \\
{webspam}   &  0.99 &  0.93 &     0.8 \\
\bottomrule
\end{tabular}

    \end{minipage}
    \quad
    \begin{minipage}{.55\linewidth}
        \caption{The first three columns list the median adversarial distance $\delta_{\mathrm{med}}$ (also called \textit{empirical robustness}) for each ensemble type, averaged over 5 folds. The last column lists the attack model $\epsilon$ for the GROOT ensembles.}%
        \label{tab:median_delta}
        \centering
        \footnotesize
        \begin{tabular}{llS[table-format=2.3]S[table-format=2.3]S[table-format=2.3]S[table-format=2.3]S[table-format=2.3]S[table-format=2.3]S[table-format=2.3]S[table-format=2.3]}
\toprule
& \multicolumn{3}{c}{$\delta_\mathrm{med}$} & 
{Parameter} \\
\cmidrule(r){2-4} \cmidrule(l){5-5}
{} &   {XGB} &    {RF} & {GROOT} & {GROOT $\epsilon$} \\
\midrule
{phoneme}   &   0.032 &   0.056 &    0.11 &    0.05 \\
{spambase}  &  0.0021 &  0.0026 &    0.05 &    0.05 \\
{covtype}   &   0.014 &    0.08 &     0.5 &    0.08 \\
{higgs}     &  0.0051 &   0.012 &    0.16 &    0.04 \\
{ijcnn1}    &   0.017 &   0.036 &    0.32 &    0.04 \\
{mnist2v4}  &   0.032 &  0.0098 &     0.4 &     0.4 \\
{fmnist2v4} &   0.016 &    0.04 &   0.018 &     0.4 \\
{webspam}   &  0.0014 &  0.0041 &   0.083 &    0.04 \\
\bottomrule
\end{tabular}

    \end{minipage} 
\end{table}

\section{AUC plots for all datasets}
\label{app:auc_all}

\begin{figure}
    \centering
    \footnotesize
    \includegraphics{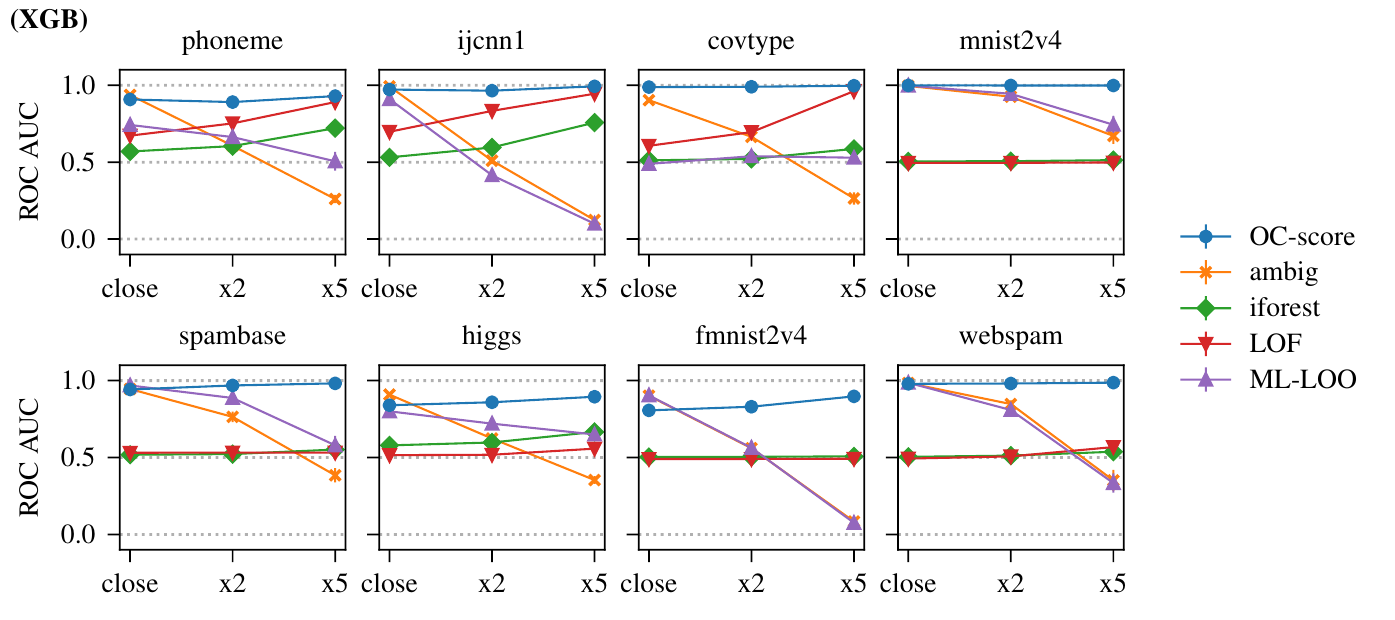}
    
    \vspace{1em}
    
    \includegraphics{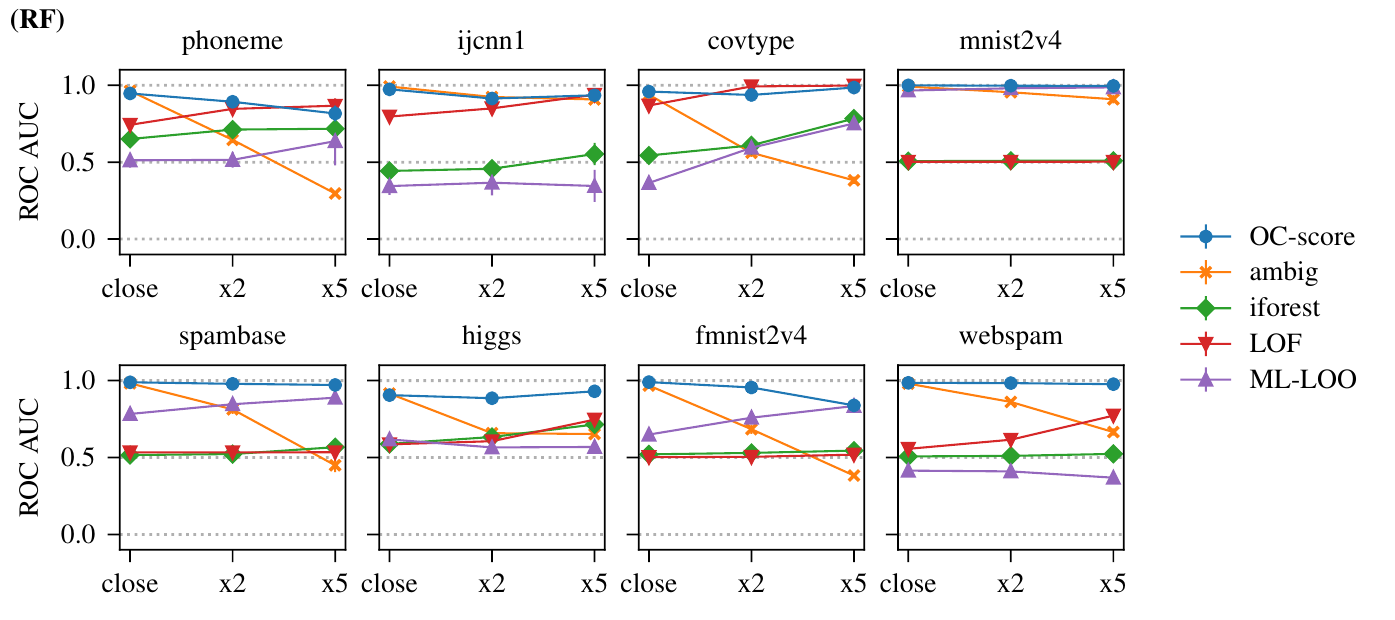}
    
    \vspace{1em}
    
    \includegraphics{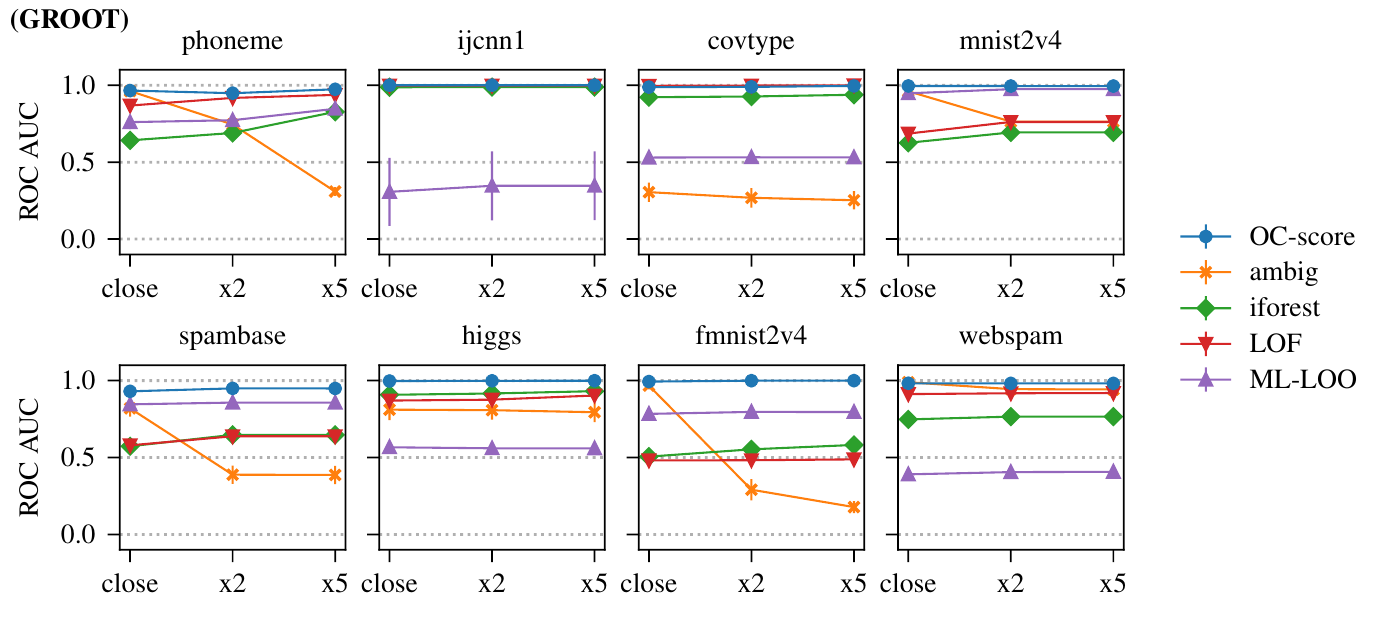}

    \caption{\textbf{Expanded version of Figure~2 in the main paper.} The average AUC ROC values for detecting adversarial vs. non-adversarial test examples as a function of the size of the adversarial perturbation for OC-score, ambiguity (\textit{ambig}), isolation forest (\textit{iforest}), local outlier factor (\textit{lof}), and ML-LOO for each of the adversarial sets. The results are shown for all eight datasets for each of the three considered tree ensemble learners. (Section~\ref{app:auc_all})}
    \label{fig:auc}
\end{figure}

The main paper only shows the ROC AUC values for the first 4 datasets in Figure~2.
Figure~\ref{fig:auc} shows ROC AUC plots for all datasets.

The ROC AUC values show how effective each method is at distinguishing adversarial examples from normal ones for a post-deployed model. Figure~\ref{fig:auc} show the performance in function of the magnitude of the adversarial perturbation.
OC-score's performance is stable across the three considered types of adversarial examples for all three ensemble types.

\section{Detection rate plots for all datasets}\label{sec:detection_rate_all}

\begin{figure}
    \centering
    \footnotesize
    \includegraphics{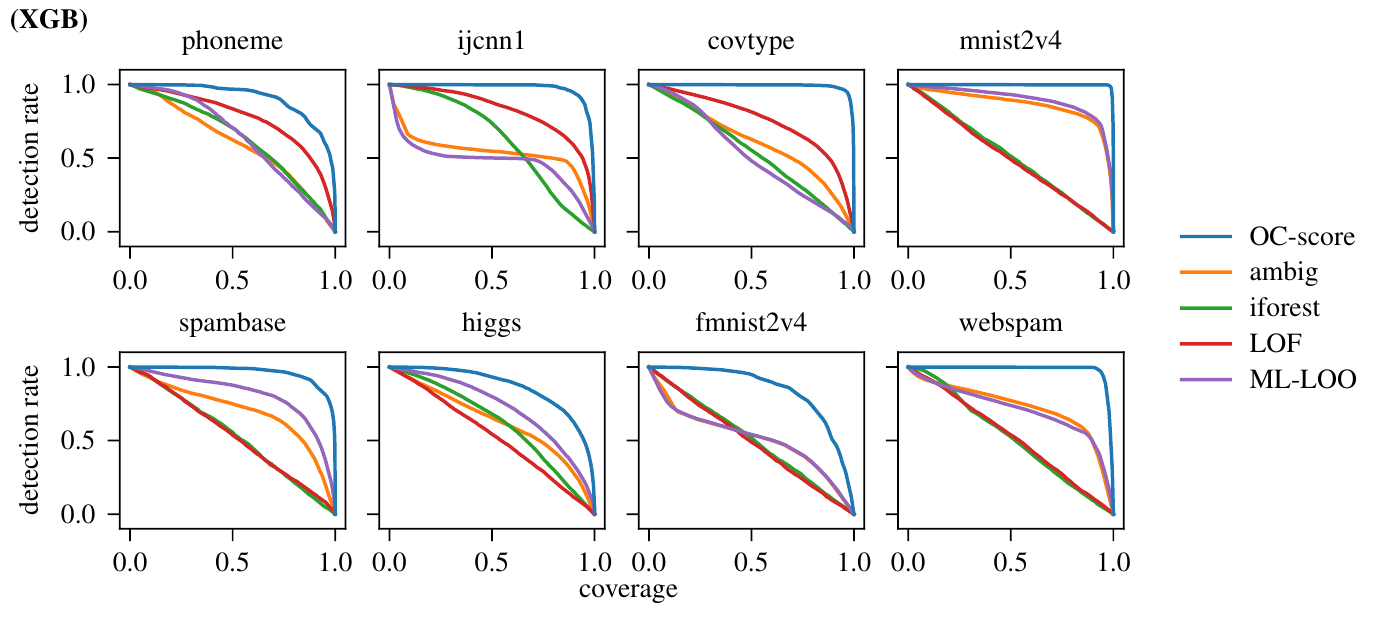}
    
    \vspace{1em}
    
    \includegraphics{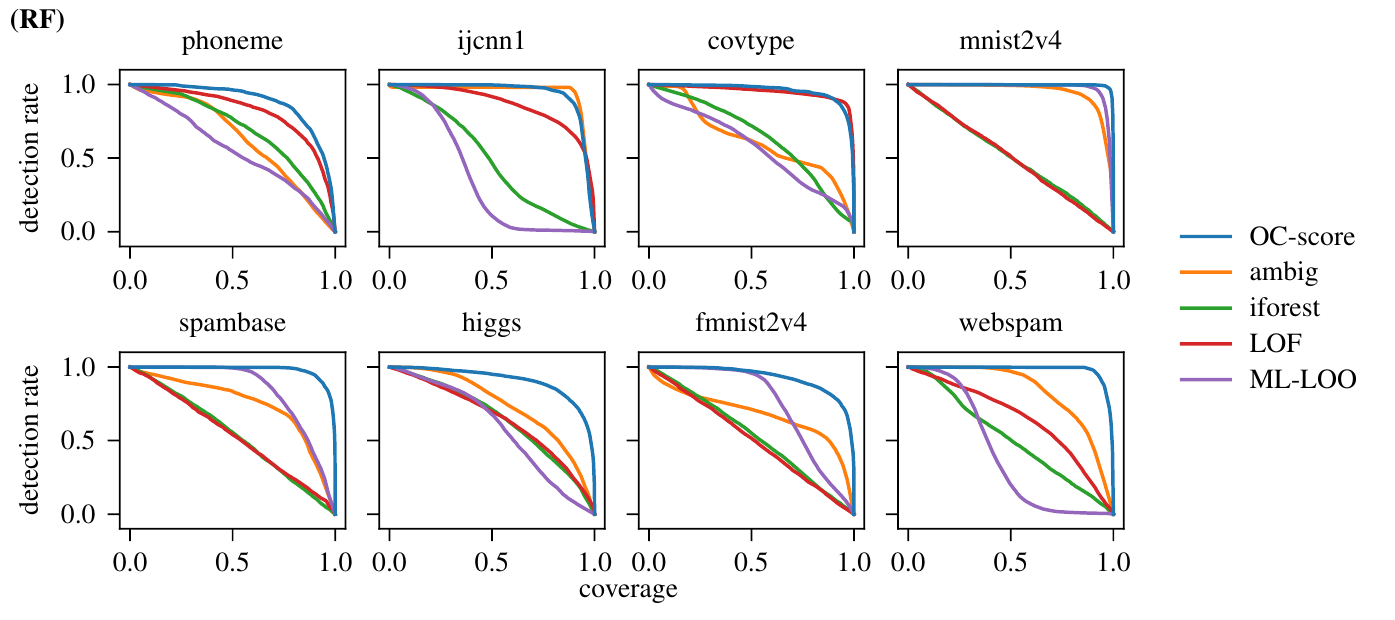}
    
    \vspace{1em}
    
    \includegraphics{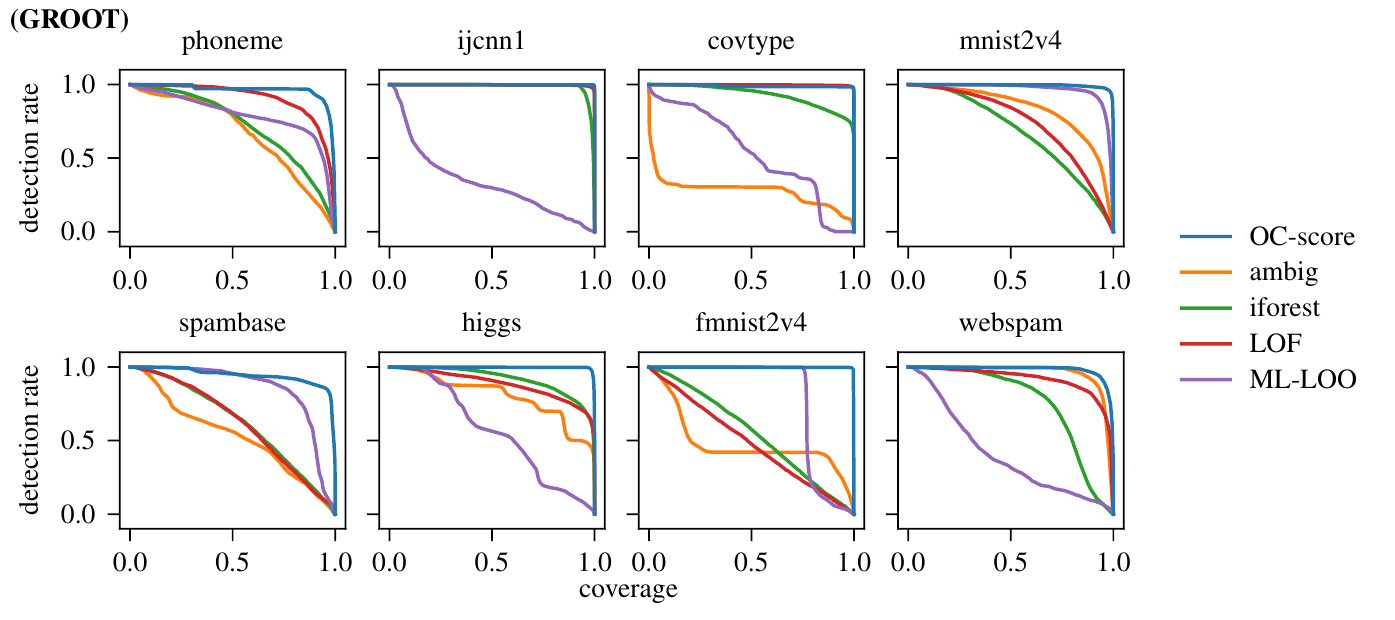}
    
    \caption{\textbf{Expanded version of Figure~3 in the main paper.} The detection rate of adversarial examples as a function of coverage for each method. The results are averaged over the three sets of adversarial examples. The higher the curve, the better.  Results are shown for XGBoost (XGB), Random Forests (RF), and GROOT. Figure 3 in the main paper only shows plots for the first four datasets. (Section~\ref{sec:detection_rate_all})}
    \label{fig:detection_rate_all}
\end{figure}

Figure~3 in the main paper only shows the coverage versus detection rate results for the first four datasets. Figure~\ref{fig:detection_rate_all} shows the coverage versus detection rate results for XGBoost, Random Forests and GROOT ensembles for all datasets. Our OC-score metric is consistently the best or equally as good as a competing method. It is also the only method that consistently performs well on all datasets and for all ensemble types.

\section{Setup time}
\label{app:setup_time}
Table~\ref{tab:setup_times} gives the setup times for each detection method. These are the one-time costs incurred during training. The setup time for OC-score is simply the time taken to encode the reference set $R$.\footnote{Note that tree learning entails sorting each training example to a leaf node, unless bagging is employed.} The setup time for \textit{iforest} and \textit{lof} is the time needed to construct the models.  \textit{lof} has by far the largest setup time because it constructs an index of the training examples. The setup times of ambiguity and ML-LOO are 0, because they do not use any auxiliary structures that need to be initialized.

\begin{table*}[]
    \centering
    \scriptsize
    \caption{The setup time in seconds for the experiment in Figure~\ref{fig:auc}. The setup time is the time to build auxiliary structures, and is averaged over 5 folds. For OC-score, the setup time is the time to encode the reference set. For \textit{iforest} and \textit{lof}, it is the time needed to construct the models. The setup time for ambiguity and ML-LOO are 0, because they do not use any auxiliary structures. (Section~\ref{app:setup_time})}
    \begin{tabular}{lS[table-format=1.4]S[table-format=1.4]S[table-format=4.4]S[table-format=3.4]S[table-format=4.4]S[table-format=1.4]S[table-format=1.4]S[table-format=4.4]}
\toprule
{} & {phoneme} & {spambase} & {covtype} & {higgs} & {ijcnn1} & {mnist2v4} & {fmnist2v4} & {webspam} \\
\midrule
{OC-score}  &   0.006 &    0.005 &       4.1 &    1.94 &    0.518 &      0.494 &       0.362 &      4.36 \\
{ambig}   &       0.0 &      0.0 &       0.0 &     0.0 &      0.0 &        0.0 &         0.0 &       0.0 \\
{iforest} &     0.107 &     0.12 &       3.9 &   0.987 &    0.514 &       1.39 &        1.42 &      10.2 \\
{LOF}     &     0.051 &    0.219 &    3310.0 &   616.0 &    204.0 &       2.53 &        2.57 &    1510.0 \\
{ML-LOO}  &       0.0 &      0.0 &       0.0 &     0.0 &      0.0 &        0.0 &         0.0 &       0.0 \\
\bottomrule
\end{tabular}

    \label{tab:setup_times}
\end{table*}

\section{Time complexity of \textit{OC-score} metric}
\label{app:time_complex}

The time complexity of the OC-score for a single example at prediction time is $O(M \cdot |D_R|)$. $M$ is the number of trees and thus the size of an output configuration. $|D_R|$ is the size of the reference set. Figure~\ref{fig:refsize_all} shows that, indeed, changing the size of the reference set has a linear effect on the evaluation time, except for the larger datasets where cache misses can have a considerable effect on the run time.
Note that this effect is CPU dependent and can be less pronounced on other CPUs.

\section{Effect of the reference set's size}
\label{app:refsize_all}

\begin{figure*}
    \centering
    \normalsize
    \textbf{(XGB)}

    \scriptsize
    \def\svgwidth{0.48\textwidth}
    \includegraphics{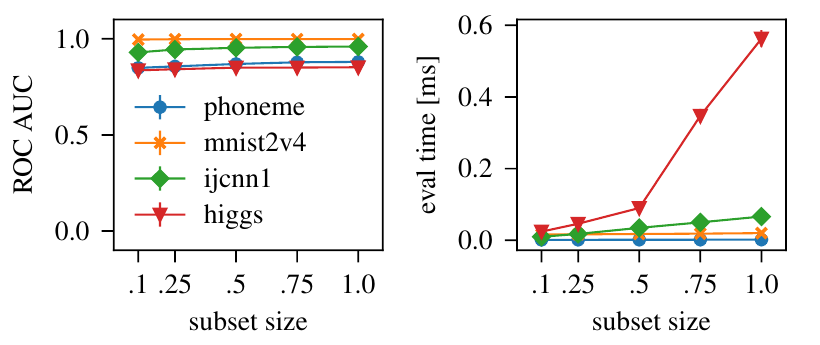}%
    \def\svgwidth{0.48\textwidth}\ \ %
    \includegraphics{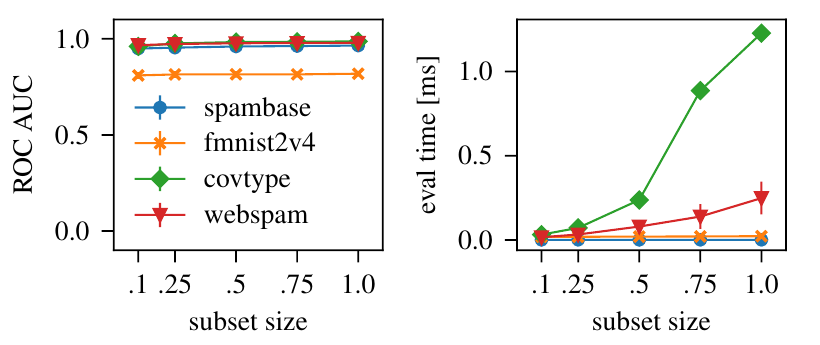}
    
    \normalsize
    \textbf{(RF)}

    \scriptsize
    \def\svgwidth{0.48\textwidth}
    \includegraphics{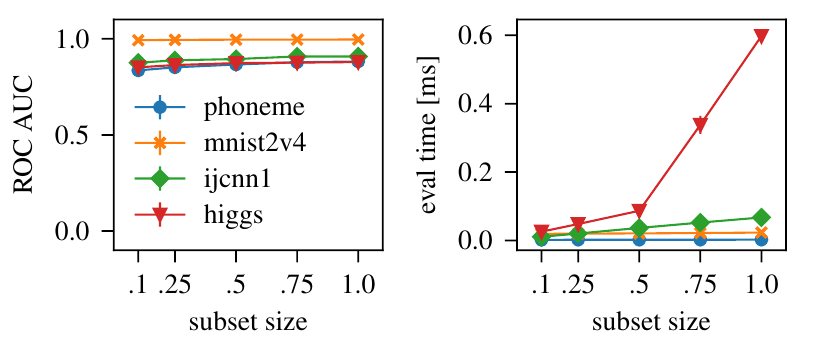}%
    \def\svgwidth{0.48\textwidth}\ \ %
    \includegraphics{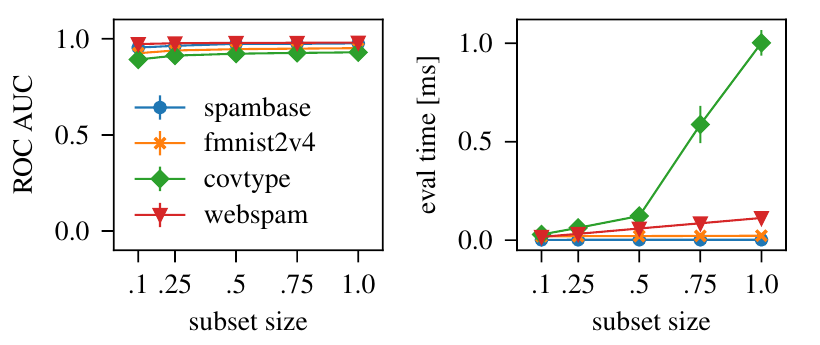}
    
    \normalsize
    \textbf{(GROOT)}
    
    \scriptsize
    \def\svgwidth{0.48\textwidth}
    \includegraphics{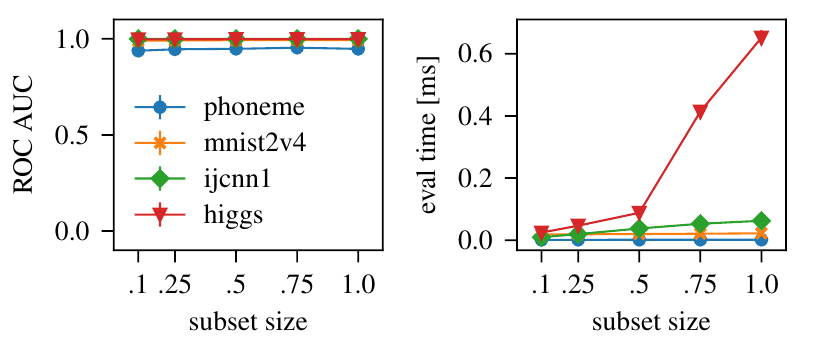}%
    \def\svgwidth{0.48\textwidth}\ \ %
    \includegraphics{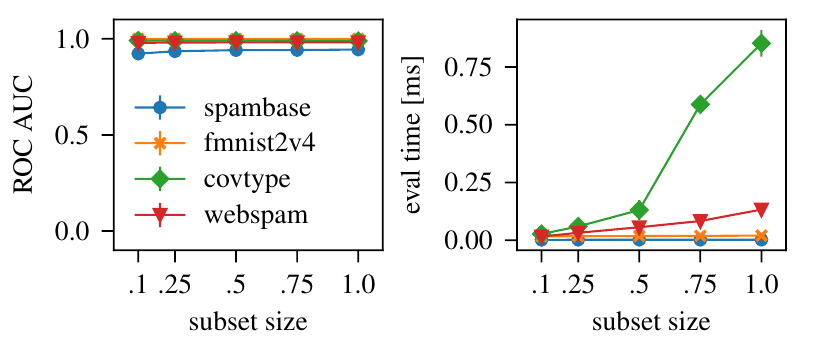}
    
    \caption{\textbf{Expanded version of Figure~4 in the main paper.} Plots in columns 1 and 3: OC-score's ROC AUC values for detecting adversarial vs. non-adversarial test examples as a function of the relative reference set size. Plots in columns 2 and 4: average time in milliseconds to compute the OC-score per example as a function of the reference set size.  Results are shown for all datasets using XGBoost (XGB), Random Forests (RF) and GROOT.  (Section~\ref{app:time_complex} \& \ref{app:refsize_all})}%
    \label{fig:refsize_all}
\end{figure*}

Figure~4 in the main paper shows the effect of varying the size of the reference set only for a selection of four datasets for XGBoost ensembles. Figure~\ref{fig:refsize_all} shows the results for all datasets and all ensemble types.
We observe similar results for all datasets and all ensemble types. The ROC AUC values (left subplots) for detecting adversarial vs. non-adversarial test examples are relatively stable, showing only a small decline in performance for the smallest reference set consisting of only 10\% of the correctly classified training set examples. The time savings (right subplots) are also consistent across datasets and ensemble types.

\section{Distribution of \textit{OC-space} hamming distance for adversarial and random perturbations}\label{sec:adv-vs-rand}

One assumption that this paper makes is that adversarial perturbations are magnified in OC-space, and that they can be picked up in OC-space using the Hamming distance. To show that this is the case, we randomly select 100 unseen correctly classified test set examples, and perturb them in two ways. The first perturbation is adversarial and flips the predicted label of the example. The second perturbation is random and has the same (small) $l_\infty$ and same $l_0$ norm (i.e., same number of affected features) as the adversarial perturbation. We repeat this 5 times, once for each fold, and average the results.

\begin{figure*}
    \footnotesize
    \centering
    
    \includegraphics{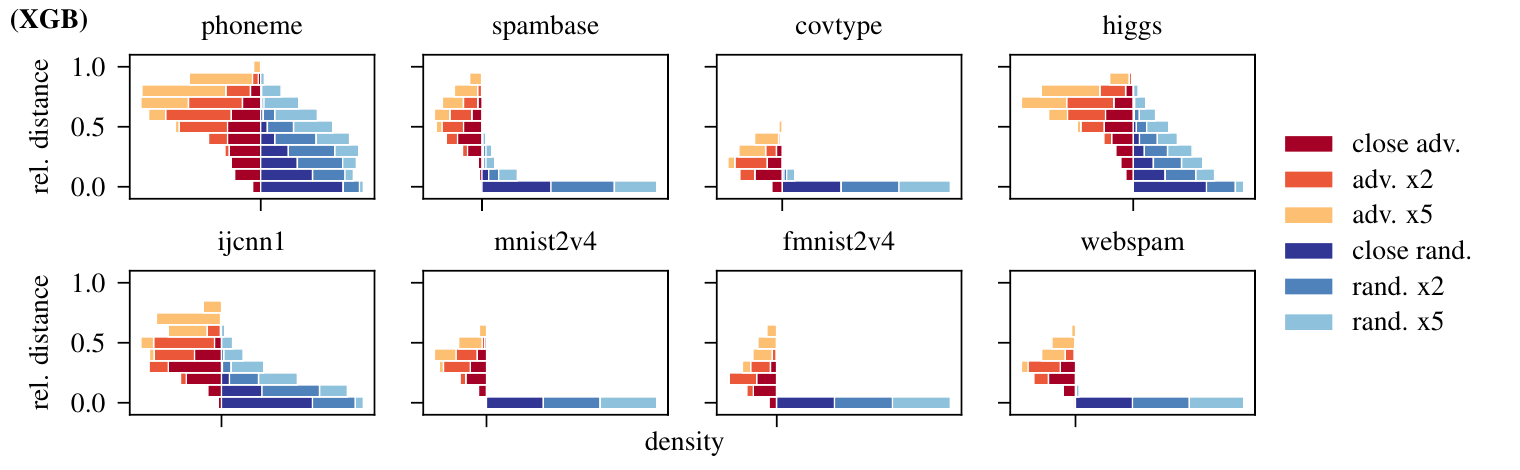}
    
    \vspace{2em}
    
    \includegraphics{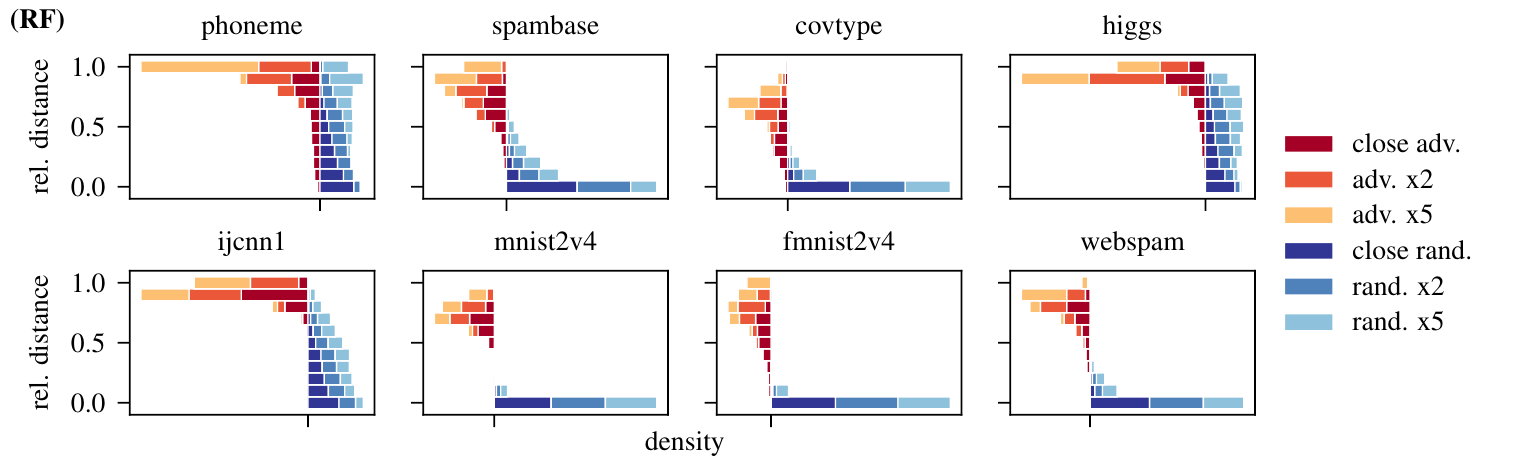}
    
    \vspace{2em}
    
    \includegraphics{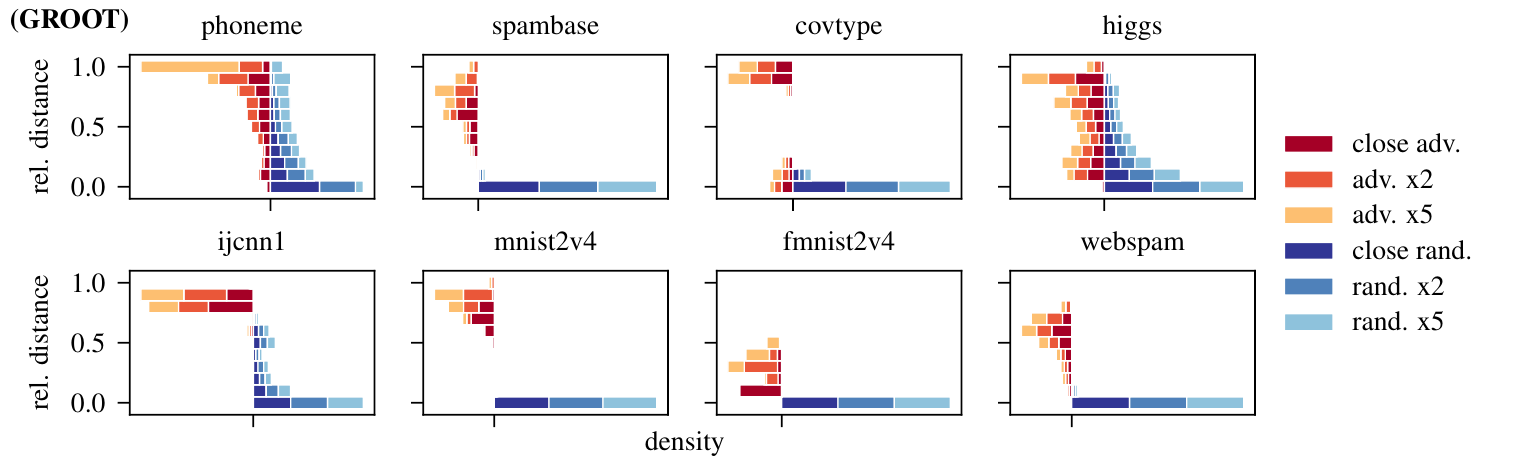}
    
    \caption{Density ($x$-axis) of Hamming distances relative to the number of trees ($y$-axis) between the original example and its adversarial perturbations (red) and random permutations (blue). A score of 1 means that all leaves in the output configuration are different. Both adversarial and random perturbations have the same $l_\infty$ and $l_0$ norms. A score of 0 means the output configuration was unaffected.  Averages over 5 folds. (Section~\ref{sec:adv-vs-rand})}
    \label{fig:adv-vs-rand}
\end{figure*}

Figure~\ref{fig:adv-vs-rand} plots the Hamming distances between the original examples and the adversarially and randomly perturbed example respectively, for the three adversarial sets and the each ensemble type.
We see that the adversarial perturbations have a much larger effect on the Hamming distance in the OC-space than the random perturbations overall, even though the magnitudes of the perturbations are the same.
This is not surprising, as in order to flip the label, the adversarial perturbation has to be carefully crafted in such a way that different leaves are activated.
This illustrates that \textbf{adversarial perturbations are magnified in OC-space}, and that a simple metric like the Hamming distance can be used to detect them.

The plots in Figure~\ref{fig:adv-vs-rand} also show that the larger the perturbations, the larger the relative distances in OC-space become. Although this effect exists for both the random and adversarial perturbations, it is more pronounced for the latter.
We also see that the smaller the number of attributes in the data, the weaker the difference is (e.g. \textit{phoneme}). With fewer attributes, a random permutation is more likely to change an attribute that is used in many trees, which in turn has an effect on the predicted leaves.

\section{Prevalence of adversarial examples}\label{sec:ratio}

While ROC analysis is generally invariant to varying class proportions, there is some debate if this is always the case~\cite{webb05}. Thus, for completeness,
we explore the effect of the prevalence of adversarial examples in the test set on the detection performance of the OC-score metric.  We reuse the same adversarial examples generated for the experiments in the main paper. To obtain the desired ratios, we randomly selected the following numbers of normal and adversarial examples respectively: (500, 500), (1000, 500), (1500, 500), (2000, 500), (2500, 500), (2400, 400), (2100, 300), (2400, 300), (1800, 200), (2500, 250). This is repeated five times for each fold, each time with a different model, different reference set, and different adversarial examples.

\begin{figure}
    \centering
    \small
    
    \textbf{(XGB)}
    
    \includegraphics{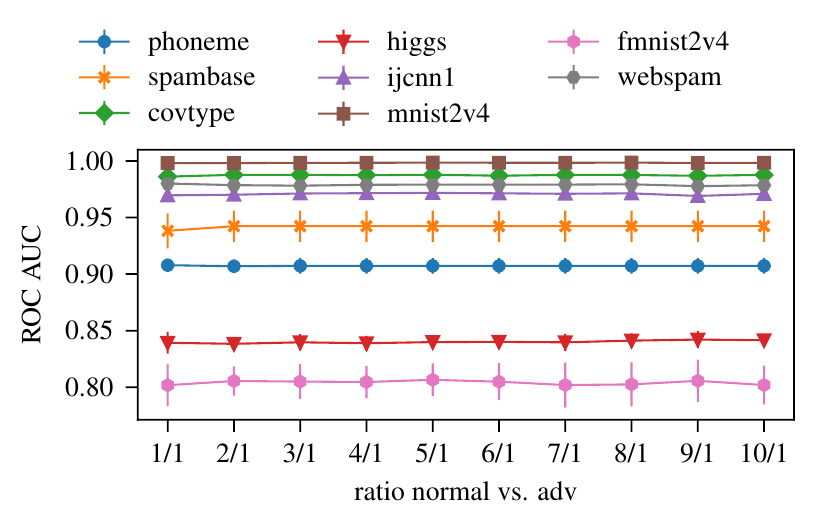}
    
    \vspace{1cm}
    \textbf{(RF)}

    \includegraphics{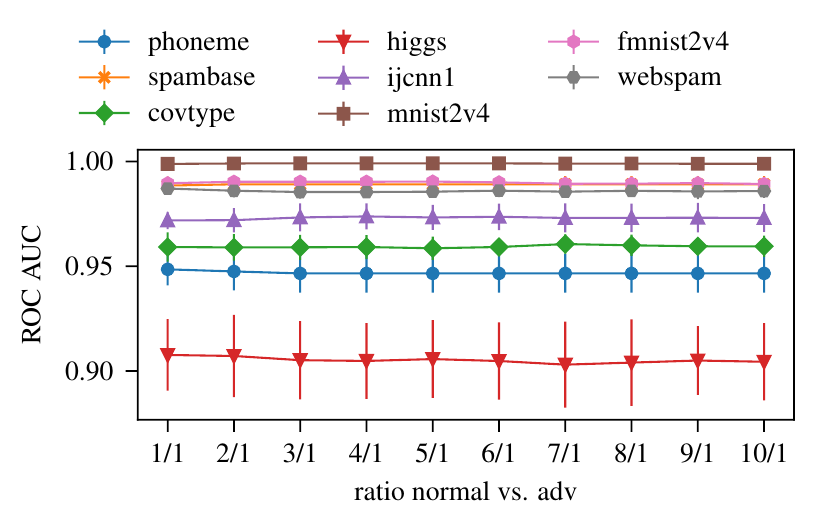}
    
    \vspace{1cm}
    \textbf{(GROOT)}
    
    \includegraphics{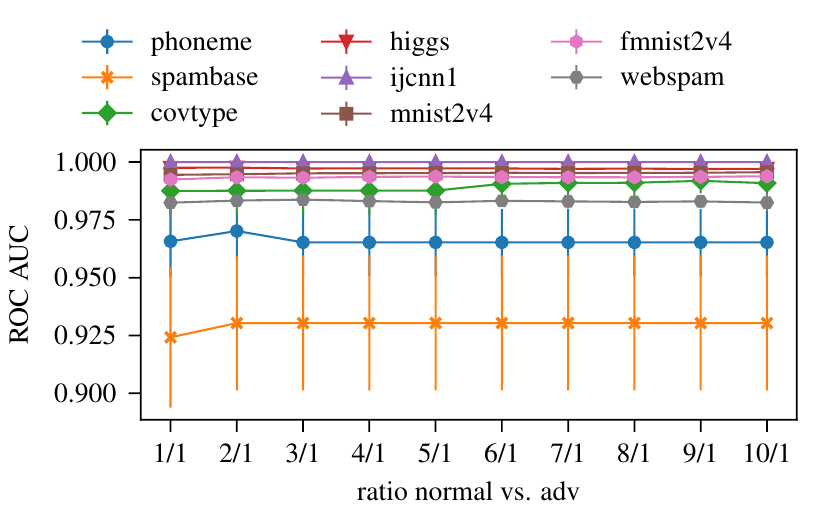}
    
    \caption{AUC ROC values for detecting adversarial vs. non-adversarial examples using OC-score as a function of the ratio of normal to adversarial examples. Results are show from XGBoost (XGB), Random Forests (RF) and GROOT (Section~\ref{sec:ratio})}
    \label{fig:ratio}
\end{figure}

Figure~\ref{fig:ratio} shows the AUC ROC values for using OC-score to detect adversarial vs. non-adversarial test examples as function of the ratio of normal to adversarial examples in the test set. Results are shown for all datasets. Regardless of the ratio, the detection performance is relatively stable on all datasets. This is true for all three considered ensemble learners.

\section{Examples of adversarial examples for each ensemble type}
\label{app:ex_adv_images}
\begin{figure}
    \centering
    \small
    \includegraphics{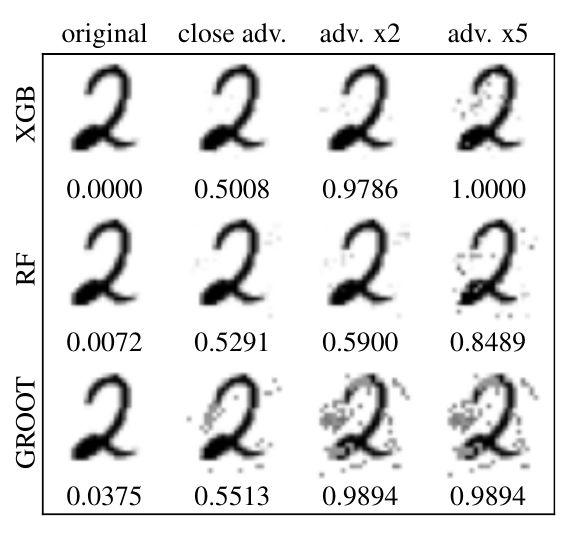}
    \includegraphics{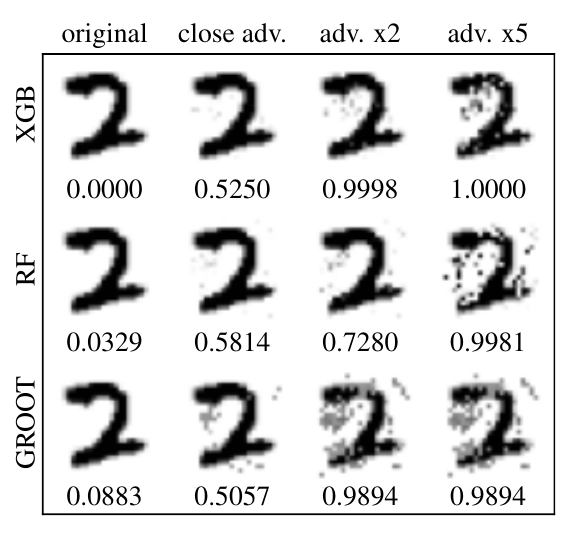}
    \includegraphics{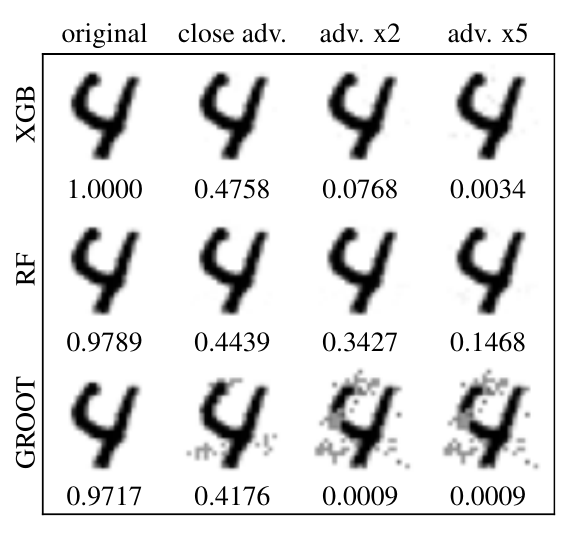}
    \includegraphics{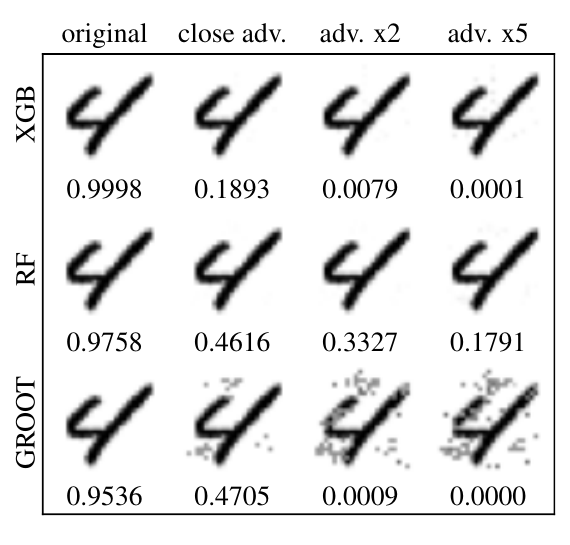}
    \caption{Examples of adversarial examples for \textit{mnist}}
    \label{fig:examples_mnist}
\end{figure}

\begin{figure}
    \centering
    \small
    \includegraphics{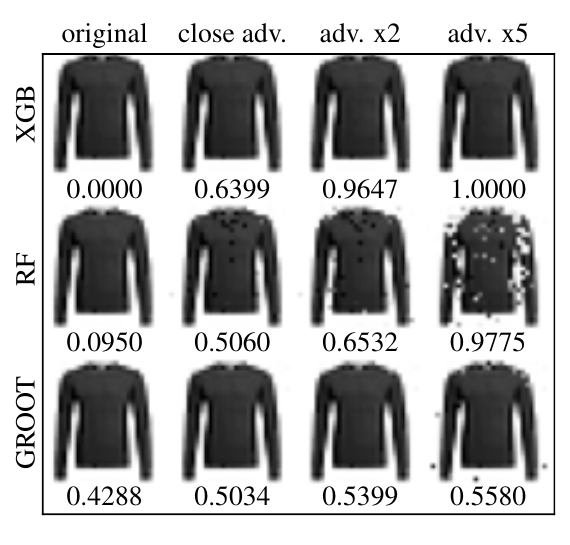}
    \includegraphics{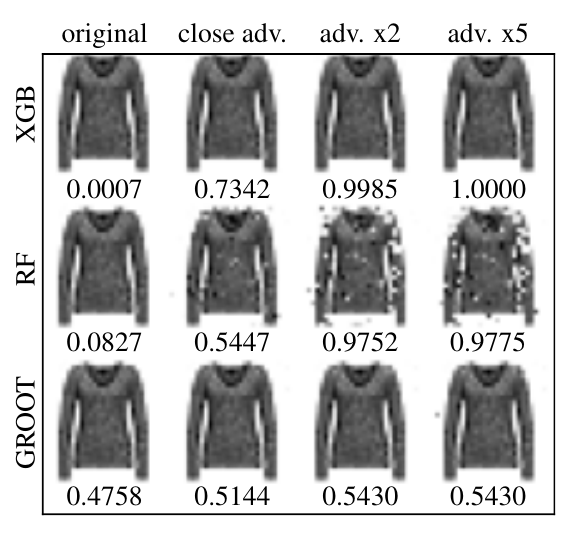}
    \includegraphics{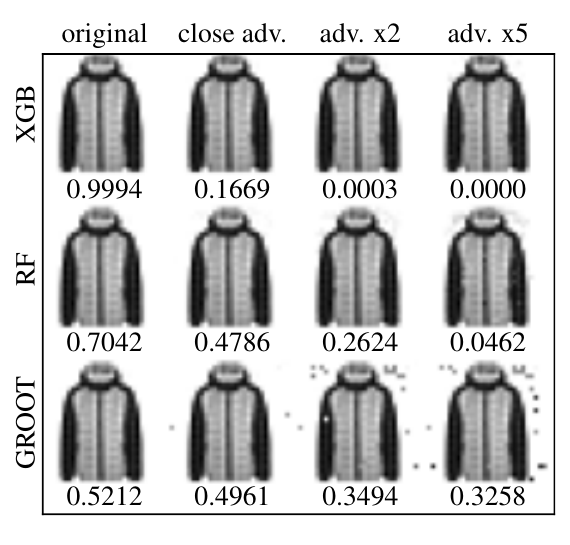}
    \includegraphics{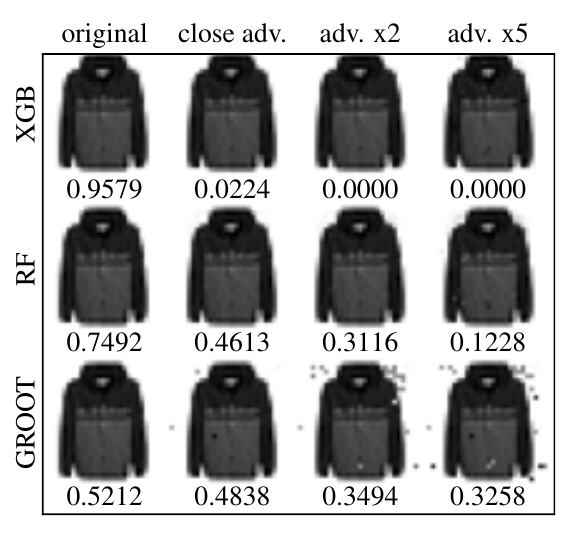}
    \caption{Examples of adversarial examples for \textit{fmnist}}
    \label{fig:examples_fmnist}
\end{figure}

Figures~\ref{fig:examples_mnist} and \ref{fig:examples_fmnist} show four adversarial examples, two of each class, for \textit{mnist} and \textit{fmnist} (pullover vs. coat). For each example and each ensemble type, we show the original example, the closest adversarial example (\textit{close adv.}), \textit{adv. x2} and \textit{adv. x5}. The predicted probability is shown below the image for each ensemble type.

A first observation is that the predicted probabilities of the closest adversarial examples are indeed close to 0.5, which explains why ambiguity tends to pick them up.

A second observation is that XGBoost is extremely confident in its predictions, and the perturbation sizes required to trick it are the smallest overall. For the 4s and all the \textit{fmnist} examples, the affected pixels are hard to discern. Moreover, XGBoost is very confident in its wrong prediction for \textit{adv. x5} examples, even though the perturbations are tiny.

A third observation is GROOT ensembles are more difficult to trick. Veritas often finds the same example for \textit{adv. x2} and \textit{adv. x5} because the space of adversarial examples is much smaller, and it thus tends to arrive at the same result even though maximum perturbation sizes are different.

\section{Error bars on coverage vs. detection rate and timings}
\label{app:checklist}

\begin{figure}
    \centering
    \footnotesize
    \includegraphics{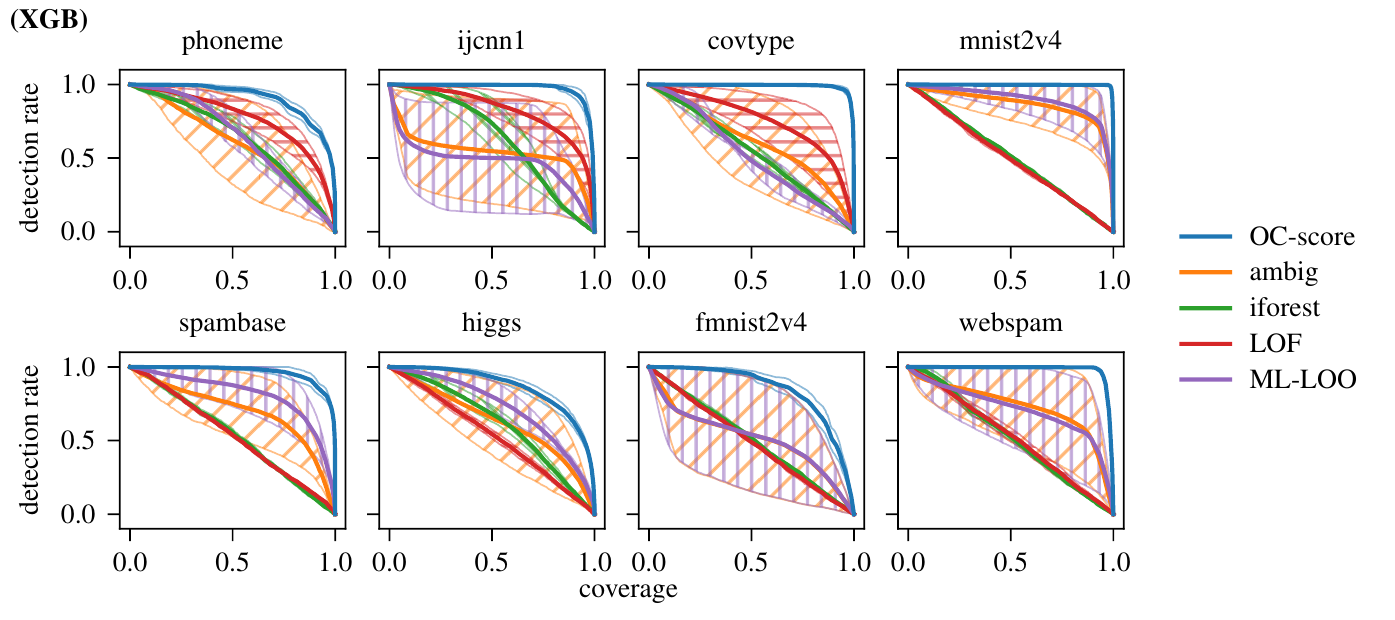}
    
    \vspace{1em}
    
    \includegraphics{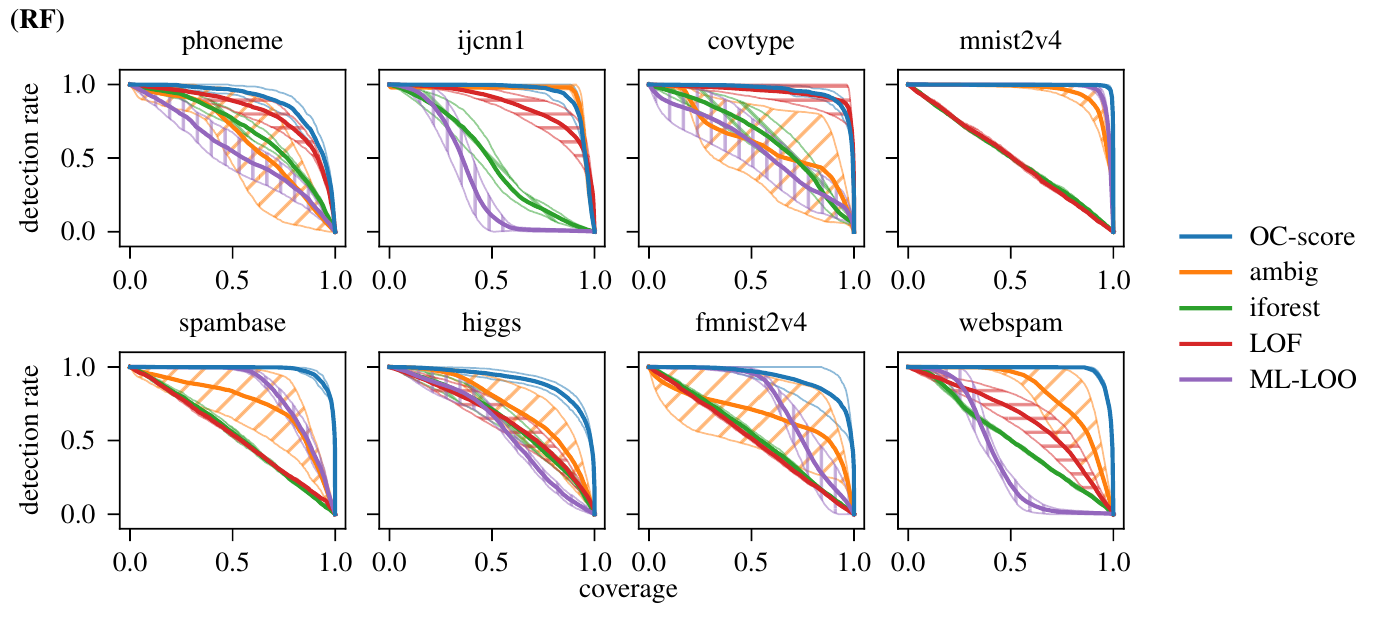}
    
    \vspace{1em}
    
    \includegraphics{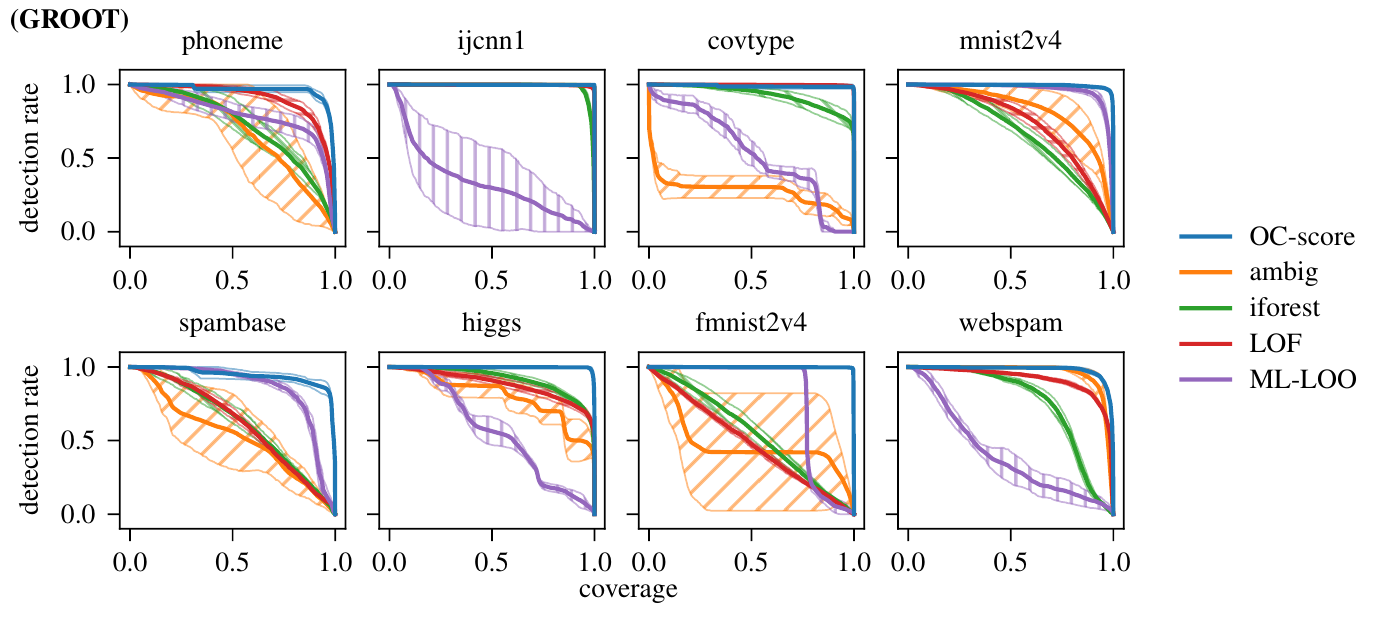}
    
    \caption{The detection rate of adversarial examples as a function of coverage for each method \textbf{with error ranges (standard deviation of detection rate)}. See Figure~\ref{fig:detection_rate_all}. These error ranges make the figures more difficult to read, but corroborate the fact that OC-score is the most reliable. (Section~\ref{app:checklist})}
    \label{fig:detection_rate_all_errorbars}
\end{figure}

We mentioned in the main paper that we did not include error ranges for the coverage versus detection rate plots for clarity. For completeness, we include these plots in Figure~\ref{fig:detection_rate_all_errorbars}. Similarly, we also include the evaluation timings table with the standard deviations in Table~\ref{tab:setup_times_std} (Table 2 in the main paper).

\begin{table*}[]
    \centering
    \scriptsize
    \caption{The evaluation times in milliseconds as in the main paper, but including the standard deviations. (Section~\ref{app:checklist})}
    \begin{tabular}{p{1.2cm}p{2.2cm}p{2.2cm}p{2.2cm}p{2.2cm}}
\toprule
{} &             {phoneme} &           {spambase} &            {covtype} &              {higgs}          \\ 
\midrule
OC-score &  0.00192 $\pm$ 0.000131 &  0.0019 $\pm$ 0.000273 &      0.973 $\pm$ 0.157 &     0.587 $\pm$ 0.0504    \\
ambig    &      0.0419 $\pm$ 0.137 &    0.0314 $\pm$ 0.0781 &    0.0198 $\pm$ 0.0428 &    0.0202 $\pm$ 0.0397    \\
iforest  &    0.0408 $\pm$ 0.00254 &   0.0461 $\pm$ 0.00425 &  0.0347 $\pm$ 0.000229 &  0.0348 $\pm$ 0.000277    \\
LOF      &    0.0133 $\pm$ 0.00288 &   0.0677 $\pm$ 0.00831 &       7.23 $\pm$ 0.168 &       3.06 $\pm$ 0.133    \\
ML-LOO   &    0.0638 $\pm$ 0.00773 &     0.103 $\pm$ 0.0133 &      0.28 $\pm$ 0.0111 &     0.331 $\pm$ 0.0157    \\
\bottomrule
\end{tabular}

\vspace{1em}

\begin{tabular}{p{1.2cm}p{2.2cm}p{2.2cm}p{2.2cm}p{2.2cm}}
\toprule
{} &             {ijcnn1} &          {mnist2v4} &         {fmnist2v4} &          {webspam} \\
\midrule
OC-score &   0.0665 $\pm$ 0.00732 &  0.0221 $\pm$ 0.00136 &  0.0218 $\pm$ 0.00165 &    0.224 $\pm$ 0.063 \\
ambig    &    0.0161 $\pm$ 0.0375 &    0.0278 $\pm$ 0.045 &   0.0278 $\pm$ 0.0455 &  0.0203 $\pm$ 0.0418 \\
iforest  &  0.0307 $\pm$ 0.000222 &   0.139 $\pm$ 0.00155 &   0.139 $\pm$ 0.00196 &  0.06 $\pm$ 0.000362 \\
LOF      &      1.84 $\pm$ 0.0229 &   0.307 $\pm$ 0.00443 &    0.308 $\pm$ 0.0253 &    5.55 $\pm$ 0.0655 \\
ML-LOO   &     0.12 $\pm$ 0.00345 &      1.54 $\pm$ 0.185 &     1.71 $\pm$ 0.0357 &   0.567 $\pm$ 0.0185 \\
\bottomrule
\end{tabular}

    \label{tab:setup_times_std}
\end{table*}

\section{Total compute time for adversarial example generation}
\label{app:compute}

Table~\ref{tab:adv_times} shows the compute time required to compute the three sets of adversarial examples for each dataset and each model type. The total time taken to compute all adversarial examples is just below 46 hours.

The total training time for the ensembles is negligible with a total training time of less than 1 hour for all models. Training a GROOT model takes on average about 10 times longer than training a scikit-learn \cite{scikit-learn} Random Forests.

\begin{table*}[]
    \centering
    \small
    \caption{The compute times for adversarial example generation in seconds. (Section~\ref{app:compute}, Table~2 in the main paper)}
    \begin{tabular}{lrrrrrrrr}
\toprule
{} &  {phoneme} &  {spambase} &  {covtype} &  {higgs} &  {ijcnn1} &  {mnist2v4} &  {fmnist2v4} &  {webspam} \\
\midrule
{xgb}   &         21 &         164 &       4726 &    19071 &      4051 &         470 &         3244 &       9775 \\
{rf}    &         37 &         406 &       3088 &    24643 &      6247 &       11752 &         9158 &       9891 \\
{groot} &         86 &         125 &      16839 &    18608 &      4948 &       10445 &         1140 &       4740 \\
\bottomrule
\end{tabular}

    \label{tab:adv_times}
\end{table*}


\end{document}